\documentclass[preprint,12pt,number]{elsarticle}

\usepackage{amssymb}
\usepackage{amsmath}
\usepackage{booktabs,amsmath}
\usepackage{booktabs,amsmath,array}  % 加上 array
\usepackage{booktabs}
\usepackage{lineno}
\usepackage{natbib}
\usepackage{threeparttable}
\setcitestyle{numbers,square} % 设置：数字编号 + 方括号
\journal{Swarm and Evolutionary Computation }

\begin{document}
%\linenumbers
%\pagewiselinenumbers

\begin{frontmatter}

%% Title, authors and addresses

%% use the tnoteref command within \title for footnotes;
%% use the tnotetext command for theassociated footnote;
%% use the fnref command within \author or \affiliation for footnotes;
%% use the fntext command for theassociated footnote;
%% use the corref command within \author for corresponding author footnotes;
%% use the cortext command for theassociated footnote;
%% use the ead command for the email address,
%% and the form \ead[url] for the home page:
%% \title{Title\tnoteref{label1}}
%% \tnotetext[label1]{}
%% \author{Name\corref{cor1}\fnref{label2}}
%% \ead{email address}
%% \ead[url]{home page}
%% \fntext[label2]{}
%% \cortext[cor1]{}
%% \affiliation{organization={},
%%            addressline={}, 
%%            city={},
%%            postcode={}, 
%%            state={},
%%            country={}}
%% \fntext[label3]{}

\title{Freezing Mechanism and Topology Dependence of Information Flow Structure in Small-Scale Particle Swarm Optimization} %% Article title

\author[add1,add2,add3]{Zhenglong Xiang\corref{cor1}}
\ead{zlxiang@nuist.edu.cn}

\author[add2]{Hongrun Wu}
%\ead{wuhongrun@whu.edu.cn}

\author[add1]{Yu Xue}
%\ead{xueyu@nuist.edu.cn}

\author[add2]{Fei Yu}
%\ead{yufei@whu.edu.cn}

\author[add4]{Yuanxiang Li}
%\ead{yxli@whu.edu.cn}

% ========== corre-author ==========
\cortext[cor1]{Corresponding author.}

\address[add1]{School of Computer Science, School of Software, \\
Nanjing University of Information Science and Technology, Nanjing 210044, China}

\address[add2]{Key Lab of Intelligent Optimization and Information Processing, \\
Minnan Normal University, Zhangzhou 363000, China}

\address[add3]{State Key Lab. for Novel Software Technology, \\
Nanjing University, Nanjing 210023, China}

\address[add4]{School of Computer Science, \\
Wuhan University, Wuhan 430072, China}

%% Abstract
\begin{abstract}
The neighborhood topology of particle swarm optimization (PSO) determines information propagation and regulates the exploration–exploitation balance. However, existing evaluation methods based on fitness and spatial diversity fail to characterize the dynamic evolution of the population's internal interaction structure. To address this gap, this paper focuses on small-scale PSO (20–50 particles) with regular topologies and models the iterative process as a dynamic weighted influence graph. We construct a multi-scale information-flow evaluation framework covering micro-, meso-, and macro-level metrics, and propose a freeze-point detection method based on effective nodes, a macro-level distributional metric. Experiments on CEC2017 benchmark functions show that the information flow structure freezes early in the search (3\%–22\% for representative configurations; 15\%–49\% for global medians) and remains unchanged thereafter. The freeze time follows the order Gbest < Von Neumann < Ring, consistent across standard, LDIW, and TVAC variants. Parameter-adaptive variants delay freezing without altering the topological order, and the freeze point is significantly negatively correlated with final solution quality. Intervention experiments under Gbest show that weakening hub nodes delays freezing and improves performance on multimodal functions, while strengthening hubs produces no additional deterioration due to saturation, revealing a decoupling between structural and performance effects. Among the three metric levels, only macro-level distributional metrics effectively capture freezing under regular topologies, because micro- and meso-level structures are predetermined by the topology. These findings reveal the freezing mechanism of information flow in small-scale PSO and provide structural prior signals for adaptive topology control.
\end{abstract}

%%Graphical abstract
%\begin{graphicalabstract}
%\includegraphics{grabs}
%\end{graphicalabstract}

%%Research highlights
%\begin{highlights}
%\item Research highlight 1
%\item Research highlight 2
%\end{highlights}

%% Keywords

\end{frontmatter}

%% Add \usepackage{lineno} before \begin{document} and uncomment 
%% following line to enable line numbers
%% \linenumbers

%% main text
%%

%% Use \section commands to start a section
\section{Introduction}
\label{sec1}
%% Labels are used to cross-reference an item using \ref command.

Particle swarm optimization (PSO), proposed by Kennedy and Eberhart in 1995 \cite{kennedy1995particle}, is one of the most representative swarm intelligence algorithms. Owing to its simple structure, few parameters, and ease of implementation, PSO has been successfully applied to numerous complex optimization problems, such as molecular simulation, inverse design of nanostructures, power grid dispatch, and graph clustering \cite{sharma2025survey,gad2022particle}. However, despite its widespread success in engineering practice, PSO has long been confronted with the inherent bottleneck of premature convergence when solving high-dimensional, multimodal, and dynamic optimization problems. Although numerous improved variants have been proposed, the intrinsic operational mechanisms of PSO, particularly the influence of the neighborhood topology on the algorithm, remain insufficiently understood \cite{nigatu2023review}.

The neighborhood topology determines the paths, speed, and range of information propagation among particles, and is a core factor affecting the balance between exploration and exploitation. The canonical PSO adopts a fully connected (Gbest) topology, which enables fast information diffusion but rapidly erodes population diversity. The ring (Ring) topology maintains diversity but often suffers from slow convergence. Kennedy et al. compared several static topologies and found that the Von Neumann topology generally achieved the best consistency, whereas the fully connected topology performed worst \cite{kennedy2002population}. Going a step further, Blackwell et al. conducted systematic experiments and confirmed this trade-off, showing that Gbest is generally the better algorithm for unimodal and separable problems, whereas a Ring neighborhood of connectivity two (Lbest) is the preferred choice for multimodal, non-separable, and composition functions \cite{blackwell2018impact}. However, these topologies are all predefined static structures that remain unchanged during the search process, and their structural influence on the search process has not yet been fully characterized. More importantly, theoretical analyses of the impact of topology on the convergence of PSO remain scarce.

However, traditional theoretical analyses of PSO have paid insufficient attention to the structural role of topology. Existing theoretical work mainly characterizes the mathematical properties of particle trajectories through stochastic convergence analysis and Lyapunov stability analysis \cite{houssein2021major}, but these analyses are typically built upon simplified assumptions, such as assuming a fully connected (Gbest) population or ignoring the influence of neighborhood topology on information propagation paths. As pointed out in a recent study \cite{deng2025collective}, many theorems assume that the expectation and variance of each particle exist and converge, which is rarely satisfied in multimodal and dynamic scenarios. More importantly, these analyses focus on the motion of individual particles rather than the population-level information interaction structure. Questions such as how topology affects information propagation, when it stops changing, and how it causally relates to search performance are difficult to answer within the traditional theoretical framework. This theoretical gap motivates the need for an alternative perspective that characterizes the structural dynamics of PSO populations during actual algorithm runs.

In recent years, network science has emerged as a promising lens for analyzing PSO dynamics. Oliveira et al. made pioneering contributions by proposing the concept of a swarm influence graph \cite{oliveira2013assessing}. They further introduced the weighted influence graph by aggregating interaction records over a sliding window \cite{oliveira2015using}, and used network metrics such as the R-value and the number of zero eigenvalues of the Laplacian matrix to diagnose stagnation \cite{oliveira2016communication}. More recently, Deng et al. \cite{deng2025collective} systematically examined the intrinsic topology of PSO and found that PSO populations inherently exhibit a small-world architecture and a heavy-tailed degree distribution. The heavy-tailed property confirms the existence of hub nodes that dominate information propagation. Despite these advances, existing studies have either provided modeling tools and a posteriori diagnostics or described static topological properties, whereas the dynamic evolution of information flow structures and the point at which this evolution ceases to change have not been systematically characterized.

Motivated by this gap, this paper investigates the freezing mechanism and topology dependence of information flow structure in small-scale PSO with regular neighborhood topologies. Particularly, we focus on the typical population size of PSO (20–50 particles), three regular topologies (Gbest, Ring, and Von Neumann), and three conventional PSO variants (standard, LDIW \cite{shi1999empirical}, and TVAC \cite{ratnaweera2004self}). The scope is deliberately restricted to this regime for three reasons. First, 20–50 particles is the most common population size in practical PSO implementations, making this regime directly relevant to real-world applications. Second, regular topologies provide a controlled experimental setting: node degrees, community structure, and path lengths are all predetermined, and the only varying factor is the distribution of edge weights. This allows us to isolate information flow concentration as a single variable and clearly observe its evolution and freezing. Third, conventional variants allow us to examine the freezing phenomenon without the confounding influence of sophisticated structural adaptive mechanisms.

To be specific, this paper establishes a dynamic network-based framework to analyze the evolution of information flow in small-scale PSO. More precisely, the PSO population is modeled as a dynamic weighted influence graph, in which particles are represented as nodes and interaction weights characterize the influence strength among particles. Based on this representation, we construct a multi-scale information-flow evaluation framework comprising micro-, meso-, and macro-level metrics. The micro-level and meso-level structures are largely determined by the predefined regular topology, whereas the macro-level metrics describe the overall distribution of influence and thus can reflect structural changes during the search. Among these metrics, we construct the effective number of nodes, Strength Gini, and Strength Entropy to characterize the macro-level distribution of information flow. On this basis, we propose a freeze-point detection method that defines the freeze point as the first sample at which the effective number of nodes series remains statistically constant over a sliding window, thereby providing a quantitative and operational criterion for structural stagnation. 

Subsequently, systematic experiments are conducted on the CEC2017 benchmark functions \cite{wu2017problem} with small-scale swarms of 20 to 50 particles and three regular topologies, namely Gbest, Von Neumann, and Ring. Three PSO variants are further considered, including standard PSO, LDIW, and TVAC. These experiments investigate when the information flow structure freezes, how the freeze time depends on topology and algorithmic variant, and how the freeze point relates to final solution quality. Notably, we further conduct intervention experiments under the Gbest topology to examine the causal effects of hub nodes by weakening and strengthening them.

The results show that the information flow structure becomes fully frozen at an early stage of the search and remains unchanged thereafter. The freeze time is earliest under Gbest, followed by Von Neumann, and latest under Ring. This order is consistent across the three PSO variants, while the parameter-adaptive variants LDIW and TVAC only delay freezing without altering the topological order. The freeze point is significantly negatively correlated with final solution quality, indicating that earlier freezing leads to worse performance. Our intervention experiments further show that weakening hub nodes delays freezing and significantly improves performance on multimodal functions, whereas strengthening hubs produces no additional deterioration due to the saturation effect, revealing a decoupling between structural effects and performance effects. These results establish a causal chain from topology to freezing and then to performance. 

We also recognize that, within the multi-scale information-flow evaluation framework developed in this study, only macro-level distributional metrics can effectively characterize the freezing phenomenon under regular topologies, because the micro-level and meso-level structures are predetermined by the topology and remain static. This indicates that macro-level metrics are more discriminative for dynamic network analysis under regular topologies.

\begin{figure}[h]
\centering
\includegraphics[width=0.9\textwidth]{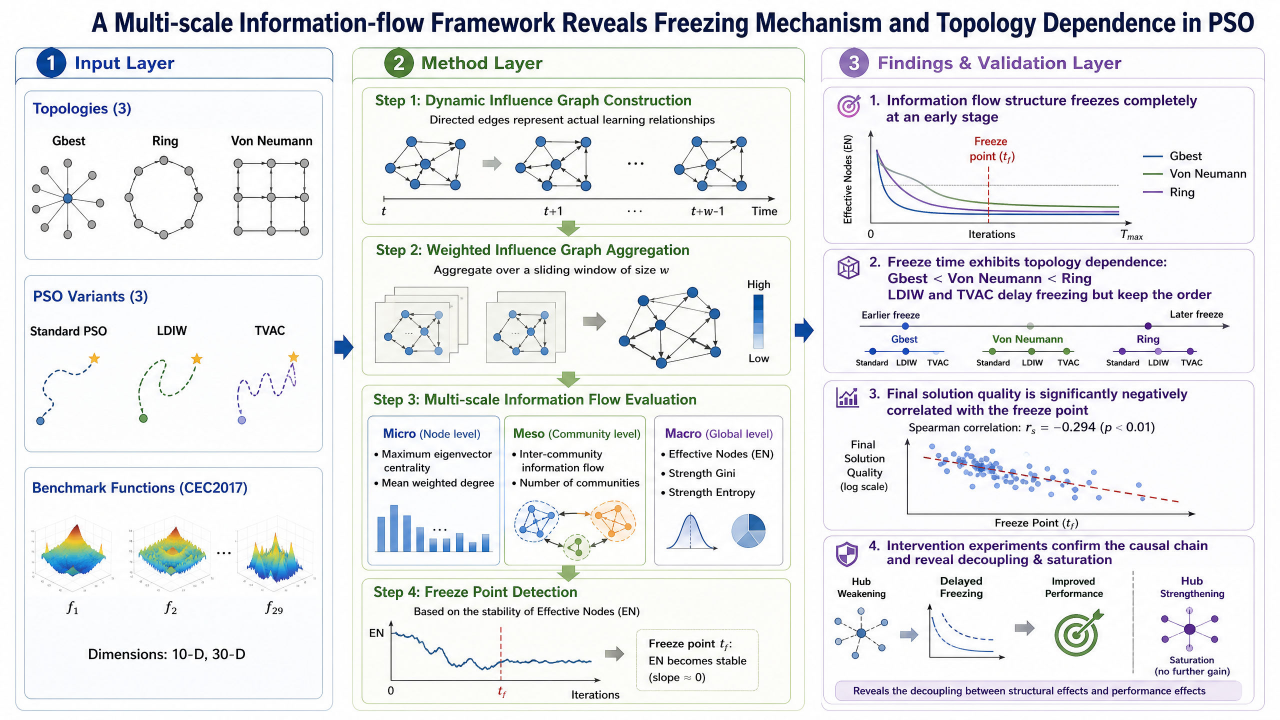} 
\caption{Overview of the proposed multi-scale information flow framework. The input layer and method layer are detailed in Section 3, and the findings and validation layer is presented in Section 4.}
\label{fig:1}
\end{figure}

The overall framework of this work is illustrated in Figure \ref{fig:1}. It consists of three layers: the input layer (topologies, variants, and benchmark functions), the method layer (dynamic influence graph construction, multi-scale information flow evaluation, and freeze-point detection), and the findings and validation layer (freezing phenomenon, topology dependence, causal validation). The detailed methodology is presented in Section 3, and the experimental results are reported in Section 4.

The main contributions are summarized as follows.
	
    (1) A multi-scale information-flow evaluation framework comprising micro-, meso-, and macro-level metrics is constructed. Within this framework, we find that the micro-level and meso-level structures are largely determined by the predefined regular topology, and that only macro-level distributional metrics can effectively reflect structural changes during the search. Among these metrics, we adopt core indicators, including the effective number of nodes, Strength Gini, and Strength Entropy, to characterize the macro-level distribution of information flow.
	
    (2) Based on the multi-scale evaluation framework, a freeze-point detection method for the PSO information flow structure is proposed. We define the freeze point as the first sample at which the effective nodes series remains statistically constant over a sliding window. This provides a quantitative and operational criterion for identifying structural stagnation.
    
	(3) The freezing phenomenon and its topology dependence are revealed. Through systematic experiments on ten CEC2017 benchmark functions, we find that the information flow structure freezes at an early stage of the search (3\%–22\% of total iterations for representative configurations; 15\%–49\% of total iterations for global medians), and the freeze time is earliest under Gbest, followed by Von Neumann, and latest under Ring. Parameter adaptive variants delay freezing but do not alter this order.
    
	(4) The correlation between the freeze point and search performance is quantified. The freeze point is significantly negatively correlated with final solution quality, so that earlier freezing leads to worse performance.

    (5) Intervention experiments provide causal evidence. Under the Gbest topology, weakening hub nodes delays freezing and significantly improves performance on multimodal functions, whereas strengthening hubs produces no additional deterioration due to the saturation effect. This reveals a decoupling between structural effects and performance effects and establishes a causal chain from topology to freezing and then to performance.

The remainder of this paper is organized as follows. Section 2 reviews related work. Section 3 presents the dynamic network modeling and multiscale information flow evaluation methodology. Section 4 reports experimental results and the freezing mechanism analysis. Section 5 discusses theoretical implications and algorithmic design insights. Section 6 concludes the paper.

%% Use \subsection commands to start a subsection.
\section{Related Work}
\label{sec2}

\subsection{PSO Neighborhood Topologies and Convergence Analysis}
The neighborhood topology of PSO determines the paths, speed, and range of information propagation among particles, directly affecting the balance between exploration and exploitation. Kennedy and Mendes \cite{kennedy2002population} compared several static topologies, including fully connected, pyramid, Von Neumann, ring, and star structures, and found that the Von Neumann topology generally achieved the best consistency, whereas the fully connected topology performed worst. Blackwell et al. \cite{blackwell2018impact} conducted systematic experiments on 84 benchmark functions and further found that Gbest is the overall better algorithm for unimodal and separable problems, whereas a ring neighborhood of connectivity two is the preferred choice for multimodal, non-separable, and composition functions. In dynamic neighborhood strategies, Liang et al. divided the swarm into random subsets that are periodically reconstructed \cite{liang2005dynamic}; Janson et al. proposed a dynamic hierarchical scheme \cite{janson2005hierarchical}. More recent efforts such as T-ADMPSO \cite{guo2026adaptive} and MSTPSO \cite{hao2025q} further explored topology-based clustering and Q-learning-driven adaptive topology switching. Valencia-Rodríguez et al. systematically analyzed the influence of the number of connections between particles on MOPSO performance using random regular graphs as swarm topologies, showing that a higher connection degree produces instability in the algorithms \cite{valencia2023influence}. These studies demonstrate that topology has a decisive impact on PSO performance.

However, traditional theoretical analyses of PSO have paid insufficient attention to the structural role of topology. Existing theoretical work mainly characterizes the mathematical properties of particle trajectories through stochastic convergence analysis and Lyapunov stability analysis \cite{houssein2021major}. Clerc et al. \cite{clerc2002particle} analyzed the explosion and stability issues of PSO from a dynamical perspective, introducing the constriction factor and proving the mathematical conditions under which particles converge. This work established the theoretical foundation for PSO convergence analysis. However, these analyses are typically based on simplified assumptions, such as assuming a fully connected population or neglecting the influence of neighborhood topology on information propagation paths. As pointed out in a recent study \cite{deng2025collective}, many theorems assume that the expectation and variance of each particle exist and converge, which is rarely satisfied in multimodal and dynamic scenarios. More importantly, these analyses focus on the motion of individual particles rather than the population-level information interaction structure. Questions such as how topology affects information propagation, when the information flow structure stops changing, and whether there is a causal link between topology and search performance are difficult to answer within the traditional theoretical framework. This theoretical gap motivates the need for an alternative perspective that characterizes the structural dynamics of PSO populations during actual algorithm runs.

\subsection{Network Science Perspective on PSO Information Flow}
To address the theoretical gap identified in Section 2.1, network science offers new tools for characterizing structural dynamics at the population level. Network science has emerged as a promising lens for analyzing PSO dynamics by treating particles as nodes and their information-exchange relationships as edges. Oliveira et al. made pioneering contributions by proposing the concept of a swarm influence graph, in which a directed edge from particle \(j\) to particle \(i\) is created if the personal best of \(j\) is used as a learning exemplar by \(i\) during a given iteration \cite{oliveira2013assessing}. This influence graph effectively captures the actual information flow rather than the static communication topology. They also introduced the weighted influence graph by aggregating interaction records over a sliding window, which suppresses random fluctuations and provides a quantitative measure of influence strength \cite{oliveira2014towards}. Based on this framework, they used network metrics such as the R-value, the number of zero eigenvalues of the Laplacian matrix, and the spectral density to diagnose stagnation \cite{oliveira2015using}. They also defined swarm communication diversity based on the connected components of the influence graph, and found that exploration is characterized by a slow differentiation of information flow, whereas exploitation relying on a few individuals leads to rapid network disruption \cite{oliveira2016communication}. Gupta et al. \cite{gupta2025enhancing} used an adapted IOHxplainer framework to systematically analyze how Ring, Star, and Von Neumann topologies affect PSO convergence and search behaviors, revealing through visualization and statistical analysis how these topologies influence information flow, diversity, and convergence speed. They found that the Von Neumann topology enhances diversity and promotes gradual convergence. This work provides explainability-level evidence that “topology determines search behavior.”

More recently, Deng et al. \cite{deng2025collective} systematically examined the intrinsic topology of PSO and found that PSO populations inherently exhibit a small-world architecture and a heavy-tailed degree distribution, with the Weibull distribution providing the best fit. The heavy-tailed property confirms the existence of hub nodes that dominate information propagation. Their small-world structure facilitates efficient cross-node information flow but simultaneously accelerates local solution assimilation and the loss of population diversity. This work provides a landmark static description of PSO's inherent topology.

Despite these advances, existing network-science studies on PSO suffer from three major limitations. First, the evolution of information flow structure has not been systematically characterized. Prior work either provides modeling tools and post-hoc diagnostics \cite{oliveira2013assessing,oliveira2016communication} or describes static topological properties \cite{deng2025collective}. Neither answers the core question of how the information flow structure evolves during the search and when it ceases to change. Second, the topology dependence of structural evolution is unclear. Different neighborhood topologies may lead to different structural evolution patterns, yet this has not been systematically compared. Third, causal evidence linking structural evolution to search performance is missing. Correlation-based analyses cannot establish whether structural changes drive performance degradation or merely accompany it. In particular, the point at which the information flow structure ceases to change, namely the structural freezing phenomenon, has not been systematically characterized; nor has how topology and algorithmic variants affect the freeze time been systematically compared.

\subsection{Adaptive Topology Mechanisms}

To overcome the limitations of static topologies, various adaptive topology strategies have been proposed. Clerc et al. \cite{clerc2003tribes} proposed TRIBES, a parameter-free PSO algorithm in which the topology and swarm size evolve automatically according to search performance. A self-adaptive topologically connected PSO (SATC-PSO) was proposed by adaptively modifying its topology connectivity in different searching stages \cite{lim2018self}. Dynamic multi-swarm PSO variants such as HIDMS-PSO further explored heterogeneous sub-swarm structures with adaptive topological connections \cite{varna2020hidms}. SET-PSO generates a new topology when the algorithm is stuck and decides whether to keep it through the Metropolis criterion, achieving adaptive adjustment of dynamic topology \cite{yin2025self}. HPSO introduced a hypergraph-based PSO topology with an adaptive hypergraph update strategy that periodically reconstructs the topology based on cumulative average particle displacement \cite{pei2026hpso}.

Recent studies have further explored the design of adaptive topologies. Wu et al. proposed SpadePSO, which adopts the Surprisingly Popular Algorithm as a complementary metric to fitness and designs a Euclidean distance-based adaptive topology where each particle connects only to its nearest neighbors \cite{wu2023incorporating}. Vafashoar et al. proposed Bifurcated PSO, which divides particles into even and uneven groups for global exploration and local fine-grained search, respectively, and employs a tree structure for dynamic neighborhood adjustment\cite{vafashoar2022bifurcated}. In high-dimensional optimization scenarios, Yang et al. \cite{yang2023random} proposed a random contrastive interaction (RCI) strategy, which randomly selects several peers for each particle to construct an interaction topology. 

These studies demonstrate that adaptive topology adjustment is effective, but most adjustment rules still rely on manually designed heuristic strategies or predefined triggering conditions, lacking mechanistic understanding of the population structural state as guidance. How to provide a structural-level triggering signal that indicates when the information flow structure has ceased to evolve and requires intervention remains a key gap.

\subsection{Search-State Diagnosis and Stagnation Detection}
Premature convergence and search stagnation are long-standing bottlenecks in PSO. In traditional methods, population spatial diversity is the most commonly used state-monitoring indicator, but its decline may reflect either the decay of exploration capability or simply the normal convergence process. Clerc et al. \cite{clerc2006stagnation} conducted a systematic mathematical analysis of particle behavior during stagnation, deriving trajectory properties under simplified models and proposing parameter improvements. However, such methods use fitness change as the sole criterion, detecting the “external manifestation” of the search rather than the evolution of the population’s internal interaction structure. Oliveira et al. pointed out that communication diversity may better reflect the essence of premature stagnation than spatial diversity, because it directly considers the social interaction relationships among particles rather than the attributes of individual particles \cite{oliveira2016communication}.

Fitness-based stagnation detection is another commonly used approach, which declares stagnation when the objective function shows no improvement over a predefined number of consecutive iterations. ADPSO introduced stagnation detection and spatial exclusion strategies in a dynamic multi-swarm PSO framework, generating “vitality particles” when subgroups fail to improve \cite{yang2023adaptive}. Adaptive diversity-driven multi-strategy PSO incorporates a multi-level parameter regulation mechanism driven by real-time diversity ratios and a stagnation penalty strategy \cite{ahmed2014efficient}. These methods are simple and computationally efficient, yet they suffer from two limitations. First, threshold setting is experience-dependent and difficult to adapt to different fitness landscapes. Second, detection can only occur after stagnation has already happened, providing no proactive early-warning signal. Clerc et al. \cite{clerc2006confinements} systematically analyzed confinement strategies and center bias in PSO, revealing sources of bias that are easily overlooked. This work demonstrates that fine-grained analysis of PSO’s intrinsic behavior beyond performance metrics is of significant value.

In terms of network science-based diagnosis, the R-value, the number of zero eigenvalues of the Laplacian matrix, and spectral density proposed by Oliveira-Júnior et al. have been used to evaluate particle search behavior and diagnose stagnation \cite{oliveira2013assessing}. The number of connected components of the swarm influence graph has been used to define communication diversity, which is correlated with population stagnation under dynamic topologies \cite{oliveira2016communication}. More recently, Deng pointed out that the heavy-tailed degree distribution and the declining trend of degree distribution entropy in PSO population communication networks may signal intensifying information aggregation \cite{deng2025collective}. However, existing research on search-state diagnosis still suffers from several deficiencies. First, the quantitative mapping between network metrics and search states has not been systematically established. Second, the early-warning capability remains insufficient, as most methods can only detect stagnation after it has occurred. Third, the multi-scale characteristics of information flow have not been systematically examined, since most works focus on single-scale network metrics.

\subsection{Summary of Research Gaps}
In summary, existing studies exhibit four key deficiencies. First, the dynamic evolution of information flow structures, particularly the freezing phenomenon, has not been systematically characterized. %Existing network analyses of PSO mainly focus on static topological properties or a posteriori diagnostic, and thus fail to assess when and why the information flow structure ceases to change. 
Second, the topology dependence and variant dependence of the freeze time remain unclear. %Existing work either provides modeling tools and macro-level diagnostics or describes static topological properties; neither addresses how different neighborhood topologies and algorithmic variants affect the freeze time. 
Third, causal evidence linking structural freezing to search performance is lacking. %Correlation-based analyses cannot determine whether structural freezing causes performance degradation or merely accompanies it. 
Fourth, under regular topologies, which levels of metrics can effectively characterize structural changes remains an open methodological question. %Although micro- and meso-level structures are predetermined by regular topologies, a systematic discussion of the effectiveness of metrics at different levels is still lacking.

\section{Dynamic Network Modeling and Multi-Scale Information Flow Evaluation}

\subsection{Problem Description and Modeling Motivation}
The population size of particle swarm optimization (PSO) is typically set between 20 and 50. This small-scale characteristic offers significant computational efficiency but introduces unique challenges for network-science-based structural analysis. In small-scale networks, traditional metrics such as clustering coefficient and modularity exhibit large statistical fluctuations, and modularity optimization suffers from an inherent resolution limit \cite{fortunato2007resolution}. Taken together, these factors indicate that directly applying mature network metrics to PSO populations is insufficient. Accordingly, method adaptation and metric screening tailored to small-scale populations are necessary.

In standard PSO, the velocity and position of particle iat iteration tare updated as:
\begin{equation}
\begin{aligned}
v_i^{t+1} &= w v_i^t + c_1 r_1 (\mathrm{pbest}_i - x_i^t) + c_2 r_2 (\mathrm{lbest}_i - x_i^t) \\
x_i^{t+1} &= x_i^t + v_i^{t+1},
\end{aligned}
\label{eq:pso}
\end{equation}
where $w$ is the inertia weight, $c_1$ and $c_2$ are learning factors, $r_1, r_2 \sim U(0,1)$ are random numbers, $pbest_{i}$ is the personal best of particle $i$, and $lbest_{i}$ is the best position in the neighborhood of particle $i$. The directed edges in the influence graph are constructed based on the selection of $lbest_{i}$.

Therefore, we aim to establish a dynamic network modeling methodology and a multi-scale information-flow evaluation metric system for small-scale PSO populations, providing a unified analytical framework for subsequently revealing the freezing mechanism. For modeling, we adopt the swarm influence graph and its weighted variant to capture actual information-exchange relationships among particles. For the metric system, we organize metrics into three levels, i.e., micro, meso, and macro. The micro-level answers “who dominates the information flow,” the meso-level answers “has the population formed subgroups,” and the macro-level answers “is information flow concentrated and when does it freeze.” Through sensitivity analysis, macro-level distributional metrics are identified as the core basis for freeze-point detection, while micro-level and meso-level metrics serve as auxiliary verification tools.

\subsection{ Construction of Dynamic Networks}
\subsubsection{Swarm Influence Graph}
The core issue in modeling a PSO population as a network is how to define the “edges” between particles. Although static communication topologies (e.g., the fully connected graph of Gbest, the ring graph of Ring) are easy to obtain, they cannot reflect the actual information flow among particles. In standard PSO, each particle’s motion is influenced by both its personal best position $pbest_{i}$ and a neighborhood best position $lbest_{i}$. However, rather than being influenced by all neighbors simultaneously, a particle selects a specific learning exemplar from its neighborhood, typically the personal best or the best individual in the neighborhood. This means that the actual “influence relationships” are much sparser than the communication topology and change dynamically with iterations.
The swarm influence graph proposed by Oliveira et al. \cite{oliveira2013assessing} effectively solves this problem. At each iteration t, a directed edge from particle $j$ to particle $i$ is created if the personal best of $j$ is used as a learning exemplar by $i$. Formally, let the adjacency matrix be $A^{(t)} \in \{0,1\}^{N\times N}$, whose elements are defined as follows.

\begin{equation}
A_{ij}^{(t)}=
\begin{cases}
1, & \text{if } \mathrm{pbest}_j \text{ influences particle } i \text{ at iteration } t \\
0, & \text{otherwise}.
\end{cases}
\label{eq:A_ij}
\end{equation}

To guarantee that every node has at least one incident edge and to reflect the cognitive component in the velocity update, we include a self-loop edge $i$→$i$. For the social component, we record the influence of the best neighbor. This generalized definition ensures graph connectivity and avoids the isolated-node problem that would otherwise distort network metrics.
\subsubsection{Weighted Influence Graph with Sliding Window}
A single-iteration influence graph only reflects instantaneous information exchange and is subject to randomness. To capture short-term statistical characteristics, we introduce a weighted influence graph with a sliding window.

\begin{equation}
W^{(t)}=\sum_{\tau=t-t_w+1}^{t} A^{(\tau)},
\label{eq:Wt}
\end{equation}
where $t_{w}=10$ is the window length. The weighted graph is then normalized by its maximum value and symmetrized as
\begin{equation}
\widetilde{W}=\frac{W}{\max(W)},\quad W_{\mathrm{sym}}=\big(\widetilde{W}+\widetilde{W}^\mathrm{T}\big)/2.
\label{eq:Wsym}
\end{equation}

This normalization preserves relative edge strength while avoiding excessive dilution. The symmetrization enables the application of undirected graph tools such as Laplacian spectral analysis.

\subsubsection{Adaptation for Small-Scale Populations}
For PSO populations ($N=$20-50), we adopt the following adaptations. (1) Connectivity guarantee: if the graph is disconnected, metrics are computed on the largest connected subgraph, and the number of connected components is recorded separately. (2) Window length: $t_{w}=10$ balances statistical stability and temporal resolution. (3) Weight normalization: dividing by $W_{\max}$ rather than $t_{w}$ preserves relative edge strength.

\subsection{Multi-Scale Information Flow Metrics}
There are many types of network metrics, but not all are applicable to small-scale, dynamic, weighted, and directed population influence graphs. Then we organize metrics into three scales, i.e., micro, meso, and macro. Each scale has a clear analytical objective, and a core metric is designated to answer the core question of that scale. Core metrics are given complete formulas in the main text; auxiliary metrics are standard definitions in network science; with concise definitions and literature references provided in the main text and complete formulas summarized in Table \ref{tab:metric_formulas} in Section 3.3.5.

\subsubsection{Micro-Scale: Who Dominates the Information Flow?}
Micro-scale metrics characterize the structural position of individual particles, with the core objective of identifying hub nodes, that is, particles that dominate the information flow. Then we define the core metric in micro-scale: Maximum eigenvector centrality $e_{\max}$. AS eigenvector centrality characterizes a node's global influence through the principal eigenvector of the adjacency matrix. For the symmetrized adjacency matrix ${{W}}_{sym}$ of the weighted influence graph, the eigenvalue equation is
\begin{equation}
W_{\mathrm{sym}} x=\lambda x,
\label{eq:eigen}
\end{equation}
where ${\lambda }_{\max}$ is the largest eigenvalue and ${x}={\left(x_{1},\; x_{2},\; \ldots ,\; x_{N}\right)}^{\top }$ is the corresponding principal eigenvector. The eigenvector centrality of node $i$ is $x_{i}$ (normalized so that $\parallel {x}{\parallel }_{2}=1$). The maximum eigenvector centrality is defined as:

\begin{equation}
e_{\max}=\max_{i} x_{i}.
\end{equation}

In the PSO context, an increase in $e_{\max}$ indicates that a particle is becoming the center of information aggregation. Under the Gbest topology, $e_{\max}$ approaches $1/\sqrt{2}$ (the theoretical value for a star graph) after freezing, quantitatively confirming the formation of hub dominance.

\textbf{Auxiliary metrics}: mean weighted degree s, standard deviation of weighted degree ${\sigma }_{s}$, and mean clustering coefficient ${\bar{C}}_{i}$. These metrics supplement the diagnosis of hub emergence, but due to large fluctuations in small-scale networks, they are not used as the primary basis. Their definitions follow standard network science literature \cite{mej2010networks,barrat2004architecture}.

It should be noted that under the Von Neumann topology, the influence graph may become disconnected in the later search stage, causing eigenvector centrality computation to degrade. The micro-level analysis of this topology is not pursued in this paper.

\subsubsection{Meso-Scale: Has the Population Formed Subgroups?}
Meso-scale metrics characterize the degree of subgroup differentiation, with the core objective of determining whether the population has formed information-clustered “small circles.”
Then we define the core metric in meso-scale: Inter-community information flow ${\bar{E}}_{AB}$. 

First, the Louvain algorithm \cite{blondel2008fast} is used to partition the influence graph into a set of communities $\mathcal{C} = \{C_1, C_2, \dots, C_k\}$. For any two communities $C_{a}$and $C_{b}$, the information flow strength between them is defined as follows:
\begin{equation}
E_{ab} = \frac{\displaystyle\sum_{i\in C_a}\sum_{j\in C_b} W_{ij}}{|C_a|\cdot|C_b|},
\end{equation}
where $W_{ij}$ is the edge weight of the symmetrized weighted influence graph and $\mid C_{a}\mid$ is the number of nodes in community $C_{a}$. The inter-community information flow is defined as the average information flow strength over all community pairs:
\begin{equation}
\bar{E}_{AB} = \frac{1}{k(k-1)}\sum_{a\neq b} E_{ab}.
\end{equation}

When ${\bar{E}}_{AB}$ is high, active information exchange exists between different subgroups; when ${\bar{E}}_{AB}$ approaches zero, the population splits into isolated subgroups that do not communicate. This metric directly reflects the degree of information integration in the population.

\textbf{Auxiliary metrics}: multi-resolution modularity $Q_{\gamma }$, number of communities $n_{c}$, and Gini coefficient of community sizes $G_{c}$. It should be noted that under regular topologies such as Ring and Von Neumann, the influence graph is structurally symmetric, and the community structure does not evolve during the search; therefore, these metrics are mainly used to verify the structural invariance of regular topologies. Their definitions follow standard network science literature \cite{fortunato2007resolution,blondel2008fast,arenas2008analysis}. Under the Gbest topology, modularity approaches zero after freezing (a star graph has no community structure), which itself is a valuable structural signal.

\subsubsection{Macro-Scale: Is Information Flow Concentrated? When Does It Freeze?}
Macro-scale metrics characterize the overall distribution of information flow, with the core objective of quantifying the degree of concentration and detecting the freeze point. Then we define the first core macro-scale metric: effective nodes $E_{\text{eff}}$. Defined as the inverse Simpson index of node strengths
\begin{equation}
E_{\text{eff}} = \frac{1}{\displaystyle\sum_{i=1}^{N} \left( s_i \bigg/\sum_{j=1}^{N} s_j \right)^2},
\end{equation}
where $s_{i}=\sum_{j=1}^{N} W_{ij}$ is the weighted degree (strength) of node $i$. When $E_{\text{eff}}\approx N$, information flow is uniformly distributed; when $E_{\text{eff}}\ll N$, flow concentrates on a few hub nodes. This metric directly quantifies the degree of hub dominance and serves as the basis for freeze-point detection.

The second core macro-scale metric is Strength Gini $G_{s}$. Measures the inequality of node strength distribution, defined as
\begin{equation}
G_s = \frac{\displaystyle\sum_{i=1}^{N}\sum_{j=1}^{N} |s_i - s_j|}{2N\displaystyle\sum_{i=1}^{N} s_i}.
\end{equation}

A higher Gini indicates more concentrated information flow. When all node strengths are equal, $G_{s}=0$; when all strength concentrates on one node, $G_{s}\rightarrow 1$.

The third core macro-scale metric is Strength Entropy $H_{s}$. The Shannon entropy of the node strength distribution
\begin{equation}
H_s = -\sum_{i=1}^{N} p_i \log p_i,\quad p_i = s_i \bigg/\sum_{j=1}^{N} s_j.
\end{equation}

It reflects the uniformity of information flow. The lower $H_{s}$, the more concentrated the flow.

These three core macro-level metrics quantify the non-uniformity of the information flow distribution from different perspectives. $E_{\text{eff}}$ is sensitive to the overall concentration of the distribution and is the most numerically stable in small-scale networks, and it is therefore selected as the primary metric for freeze-point detection in this study. $G_{s}$ is sensitive to inequality in the middle range of the distribution, whereas $H_{s}$ is sensitive to the tail of the distribution, that is, the emergence of a small number of hub nodes. The latter two serve as robustness verification metrics to confirm that the freezing phenomenon remains consistent across multiple distributional measures. If all three metrics exhibit consistent freezing behavior, the reliability of the freezing phenomenon is strengthened.

\textbf{Auxiliary metrics}: algebraic connectivity ${\lambda }_{2}$, spectral gap $\Delta\lambda$, degree distribution entropy $H_{k}$, global efficiency $E_{glob}$, and average path length $L$. These are used for robustness checks to verify whether the conclusions from core metrics hold under different measures. Their definitions follow standard network science literature \cite{fiedler1973algebraic,mej2010networks}.

\subsubsection{Summary of Multi-Scale Metrics}
Table \ref{tab:metrics_summary} summarizes the multi-scale information flow evaluation system. The “Role” column indicates whether a metric is core or auxiliary. Core metrics are given complete formulas in the main text; auxiliary metrics are defined in Table \ref{tab:metric_formulas} in Section 3.3.5. Among the core macro-level metrics, 
$E_{\text{eff}}$ serves as the primary metric for freeze-point detection, whereas $G_{s}$ and $H_{s}$ serve as robustness verification metrics. A complete comparison is provided in Section 4.2 and in Figure S2 and Table S2 of the Supplementary Material.

\begin{table}[htbp]
  \centering
  \small
  \caption{Multi-scale information flow metrics for PSO populations}
  \label{tab:metrics_summary}
  \begin{tabular}{l l l l l}
    \toprule
    Scale & Core Question & Metric & Symbol & Role \\
    \midrule
    Micro & Who dominates?
          & Max eigenvector centrality
          & $e_{\text{max}}$
          & \textbf{Core} \\
    Micro &
          & Mean/std of weighted degree
          & $\bar{s},\,\sigma_s$
          & Auxiliary \\
    Micro &
          & Mean clustering coefficient
          & $\bar{C}_i$
          & Auxiliary \\

    Meso  & Subgroups formed?
          & Inter‑community flow
          & $\bar{E}_{AB}$
          & \textbf{Core} \\
    Meso  &
          & Multi‑resolution modularity
          & $Q_\gamma$
          & Auxiliary \\
    Meso  &
          & Number of communities
          & $n_c$
          & Auxiliary \\
    Meso  &
          & Gini of community sizes
          & $G_c$
          & Auxiliary \\

    Macro & Concentrated? Frozen?
          & Effective Nodes
          & $E_{\text{eff}}$
          & \textbf{Core} \\
    Macro &
          & Strength Gini
          & $G_s$
          & \textbf{Core} \\
    Macro &
          & Strength Entropy
          & $H_s$
          & \textbf{Core} \\
    Macro &
          & Algebraic connectivity
          & $\lambda_2$
          & Auxiliary \\
    Macro &
          & Degree distribution entropy
          & $H_k$
          & Auxiliary \\
    Macro &
          & Global efficiency
          & $E_{\text{glob}}$
          & Auxiliary \\
    Macro &
          & Spectral gap
          & $\Delta\lambda$
          & Auxiliary \\
    Macro &
          & Average path length
          & $L$
          & Auxiliary \\
    \bottomrule
  \end{tabular}
\end{table}

\subsubsection{Summary of Metric Formulas}
Table \ref{tab:metrics_summary} summarizes the formulas for all 15 metrics. Core metrics have been detailed in Sections 3.3; auxiliary metrics are standard definitions in network science, listed here for reproducibility. In Table \ref{tab:metric_formulas}, $N$ is the population size, $W_{ij}$ is the edge weight of the symmetrized weighted influence graph, $s_{i}=\sum_{j} W_{ij}$ is the strength of node $i$, $k_{i}$ is the degree of node $i$, $d\left(i,\; j\right)$ is the shortest path length between nodes $i$ and $j$, $\boldsymbol{L}=\boldsymbol{D}-{\boldsymbol{W}}_{sym}$ is the Laplacian matrix, and $m=\frac{1}{2}\sum_{ij} W_{ij}$ is the total weight.

Table \ref{tab:metric_formulas} summarizes the formulas of all metrics, core metrics ($e_{\max}$, ${\bar{E}}_{AB}$, $E_{\text{eff}}$, $G_{s}$, $H_{s}$) have been detailed in the main text; auxiliary metric formulas are standard network science definitions, with detailed derivations in references \cite{fortunato2007resolution,mej2010networks,blondel2008fast,arenas2008analysis,fiedler1973algebraic}. 

\begin{table}[htbp]
\centering
\small
\setlength{\tabcolsep}{4pt}
\caption{Summary of metric formulas}
\label{tab:metric_formulas}
\begin{tabular}{l p{3.2cm} l >{$\displaystyle}p{5.6cm}<{$} c}
\toprule
Scale & Metric & Symbol & \text{Formula} & Ref. \\
\midrule
Micro
& Max eigenvector centrality
& $e_{\text{max}}$
& \max_i x_i,\ \text{where } \mathbf{W}_{sym}\mathbf{x} = \lambda_{\text{max}} \mathbf{x}
& This work \\
Micro
& Mean weighted degree
& $\bar{s}$
& \frac{1}{N}\sum_i s_i
& \cite{mej2010networks} \\
Micro
& Std. of weighted degree
& $\sigma_s$
& \sqrt{\frac{1}{N}\sum_i (s_i-\bar{s})^2}
& \cite{mej2010networks} \\
Micro
& Mean clustering coefficient
& $\bar{C}_i$
& \frac{1}{N'}\sum_{i:k_i\ge 2} \frac{2E_i}{k_i(k_i-1)}
& \cite{barrat2004architecture} \\
Meso
& Inter-community flow
& $\bar{E}_{AB}$
& \frac{1}{k(k-1)}\sum_{a\neq b} \frac{\displaystyle\sum_{i\in C_a}\sum_{j\in C_b} W_{ij}}{|C_a|\cdot|C_b|}
& This work \\
Meso
& Multi-resolution modularity
& $Q_\gamma$
& \frac{1}{2m}\sum_{ij}\left(W_{ij}-\gamma\frac{s_i s_j}{2m}\right)\delta(c_i,c_j)
& \cite{fortunato2007resolution} \\
Meso
& Number of communities
& $n_c$
& \text{Number of Louvain partitions}
& \cite{blondel2008fast} \\
Meso
& Gini of community sizes
& $G_c$
& \frac{1}{2 N_c \bar{c}} \sum_{a}\sum_{b} |c_a - c_b|
&  \cite{arenas2008analysis} \\
Macro
& Effective Nodes
& $E_{\text{eff}}$
& 1 \Big/ \sum_i \biggl( s_i \Big/ \sum_j s_j \biggr)^2
& This work \\
Macro
& Strength Gini
& $G_s$
& \frac{1}{2N \bar{s}} \sum_{i}\sum_{j} |s_i - s_j|
& This work \\
Macro
& Strength Entropy
& $H_s$
& -\sum_i p_i \log p_i,\; p_i = s_i \Big/ \sum_j s_j
& This work \\
Macro
& Algebraic connectivity
& $\lambda_2$
& \parbox[t]{5.4cm}{\centering Second smallest eigenvalue of Laplacian matrix}
& \cite{fiedler1973algebraic} \\
Macro
& Spectral gap
& $\Delta\lambda$
& \lambda_N - \lambda_2
& \cite{mej2010networks} \\
Macro
& Degree distribution entropy
& $H_k$
& -\sum_k p(k)\log p(k)
& \cite{fiedler1973algebraic}  \\
Macro
& Global efficiency
& $E_{\text{glob}}$
& \frac{1}{N(N-1)}\sum_{i\neq j}\frac{1}{d(i,j)}
& \cite{fiedler1973algebraic}  \\
Macro
& Average path length
& $L$
& \frac{1}{N(N-1)}\sum_{i\neq j} d(i,j)
& \cite{fiedler1973algebraic}  \\
\bottomrule
\end{tabular}
\end{table}

\subsection{Computability and Statistical Stability}
\subsubsection{Statistical Characteristics of Small-Scale Networks}
We observe that small-scale networks differ from large-scale networks in three essential aspects. First, the discretization of metric values restricts node degrees to integer values from 0 to 29, thereby limiting the possible shapes of degree distributions. Second, the modularity resolution limit implies that communities with fewer than $\sqrt{L/2}$ edges may be ignored \cite{mej2010networks}, for $N=30$ and an average degree of 6, this threshold is approximately 6.7 edges. Third, the spectral gap is unstable because eigenvalues are sensitive to small perturbations when ${\lambda }_{2}$ approaches zero.

\subsubsection{Metric Screening and Selection Logic}
Through autocorrelation analysis \cite{box2015time}, spearman correlation analysis \cite{spearman1904proof}, and redundancy analysis \cite{prichard1995generalized}, we screen core metrics from the 16 initial metrics. The selection logic is as follows: the macro-level metric $E_{\text{eff}}$ is most sensitive to search state changes and insensitive to local connectivity changes, and is therefore selected as the sole core basis for freeze-point detection. The micro-level metric $e_{\max}$ provides supplementary evidence for hub formation under Gbest and Ring topologies; under Von Neumann, due to late-stage disconnection of the influence graph, this metric degrades and is used only as auxiliary. The meso-level metric ${\bar{E}}_{AB}$ exhibits small values and limited variation under regular topologies, and is therefore not used as the primary basis for freeze-point detection. Its typical evolution curve is shown in Figure \ref{fig:f4-6} of the main text. 

Among the 16 metrics constructed, 5 are selected as core metrics ($E_{\text{eff}}$, $G_s$, $H_s$, $e_{\max}$,  ${\bar{E}}_{AB}$) and analyzed in detail in the main text. The remaining 11 auxiliary metrics are used for robustness verification and structural invariance checks under regular topologies. Figure S4 in the Supplementary Material presents the complete evolution curves of all 16 metrics, demonstrating that all metrics have been computed and analyzed. Auxiliary metrics such as weighted degree, clustering coefficient, algebraic connectivity, and degree distribution entropy are not expanded in the main text because their variation is limited under regular topologies (as explained in Section 3.3). 

The screening process further considers the sensitivity of the freeze time to algorithmic variants and topologies, ensuring that the selected core metrics remain discriminative across different configurations. Accordingly, $E_{\text{eff}}$ is designated as the primary metric for freeze-point detection, while $G_s$ and $H_s$ aare retained as robustness verification metrics within the core macro-level metric group.

\subsection{Freeze-Point Detection Method}
\subsubsection{Definition}
We define the freeze point as the first sample index $t^{\ast }$ such that
\begin{equation}
\mathrm{std}\big(E_{\mathrm{eff}}[t^{*}-W:t^{*}]\big)<\varepsilon,
\label{eq:std_eff_energy}
\end{equation}
where $W=50$ is the detection window and $\varepsilon ={10}^{-3}$ is the tolerance. The freeze point marks the transition from structural evolution to structural stagnation.

\subsubsection{Parameter Sensitivity}
We tested $W \in \{30,\,50,\,100\}$ and $\varepsilon \in \{10^{-4},\,10^{-3},\,10^{-2}\}$. Although the freeze point values exhibit some sensitivity to these parameters, especially for the Ring topology, which is more sensitive to the window length $W$ due to the slow drift of its curve, the order of freeze points across the three topologies, Gbest < Von Neumann < Ring, remains consistent under all parameter combinations. This indicates that the topology dependence of the freeze time is robust to the detection parameters. Detailed sensitivity analysis results are provided in Table S4 of the Supplementary Material.

\subsubsection{Intuitive Interpretation}

When Effective Nodes $E_{\text{eff}}$ stabilizes, it means that the distribution of information flow among particles has become fixed: the same set of hub nodes dominates the flow, and no new nodes enter or leave the effective flow structure. In other words, the learning relationships among particles have converged. The freeze point therefore provides a quantitative, operational criterion for identifying when the population has structurally stopped exploring.

\subsection{Experimental Design}
\subsubsection{Benchmark Functions and Algorithm Configuration}
Experiments use 10 representative CEC2017 functions (F1, F3, F8, F10, F15, F18, F20, F25, F28, F29) in both $10D$ and $30D$ settings. Three topologies (Gbest, Ring, Von Neumann) and three PSO variants (standard, LDIW, TVAC) are considered. Population size $N=30$, maximum iterations $D$×$500$, 15 independent runs per configuration. The three variants are included to examine whether the freezing phenomenon is sensitive to parameter-adaptive strategies. The effect of variants on freeze time is reported in Section 4.3.
\subsubsection{Data Collection and Freeze-Point Detection}
At each iteration, the influence graph adjacency matrix is recorded. The weighted influence graph is constructed with $t_{w}=10$, and metrics are computed every 10 generations. The freeze point is detected using the Effective Nodes series.

\subsubsection{Search State Calibration}
Search states are calibrated based on the freeze point: exploration (before freeze/2), exploitation (freeze/2 to freeze), and frozen (after freeze). This provides a physically meaningful stage division rather than an arbitrary time split.

This Section has established a dynamic network modeling methodology and a multi-scale information flow evaluation framework for small-scale PSO populations, along with a freeze-point detection method based on Effective Nodes. The next section will systematically validate the existence of the freezing phenomenon, its topology dependence, and its causal relationship with search performance on the CEC2017 benchmark functions.
%\subsection{Summary}
%This section establishes a dynamic network modeling methodology and a multi-scale information-flow evaluation metric system for small-scale PSO populations. The swarm influence graph and its weighted sliding-window variant are adopted to capture actual information-exchange relationships. A multi-scale metric system covering micro, meso, and macro levels is constructed, in which macro-level metrics $E_{\text{eff}}$, $G_{s}$, and $H_{s}$ are the core basis for freeze-point detection, while micro-level $e_{\max}$ and meso-level ${\bar{E}}_{AB}$ serve as auxiliary verification. A freeze-point detection method based on Effective Nodes is proposed, and the experimental design is provided. The methodology established in this section provides a unified analytical framework for the subsequent analysis of the freezing mechanism.

\section{Experimental Results and Freezing Mechanism Analysis}
\subsection{Experimental Setup}

We focuses on small-scale PSO, i.e., population sizes of 20–50 particles. In the experiments, the population size is fixed at $N=30$, and the problem dimension is set to 10 and 30 respectively to examine the freezing phenomenon under different dimensions. Experiments were conducted on 10 CEC2017 benchmark functions (F1, F3, F8, F10, F15, F18, F20, F25, F28, F29) in both $10D$ and $30D$ settings. Three PSO variants (standard, LDIW, TVAC) and three regular topologies (Gbest, Ring, Von Neumann) were considered. Population size was N=30, maximum iterations $D$×500, and 15 independent runs per configuration. Freeze-point detection used window $W=50$ and tolerance $\varepsilon ={10}^{-3}$. Table \ref{tab:exp_config} summarizes the experimental configuration.

The main text primarily focuses on the $30D$ results for analysis. The $10D$ results exhibit the same freezing trend as $30D$: identical topology order (Gbest < Von Neumann < Ring) and similar freeze proportions (both within the early 3\%–11\% range of the search). The freezing curves for representative configurations are provided in Figure S1 of the Supplementary Material, and a comparison of freeze points between $10D$ and $30D$ is given in Table S1 of the Supplementary Material. A summary of the freeze points for all 180 configurations, including  $10D$  and $30D$, is provided in Table S3 of the Supplementary Material.

\begin{table}[htbp]
  \centering
  \small
  \caption{Experimental configuration}
  \label{tab:exp_config}
  \begin{tabular}{ll}
    \toprule
    Parameter & Setting \\
    \midrule
    Benchmark        & CEC2017 (10 functions) \\
    Dimension       & 10, 30 \\
    PSO variants    & Standard, LDIW, TVAC \\
    Topologies      & Gbest, Ring, Von Neumann \\
    Population size & 30 \\
    Max iterations  & $D \times 500$ \\
    Sliding window  & 10 \\
    Independent runs& 15 \\
    Freeze detection& $W=50,\, \varepsilon=10^{-3}$ \\
    \bottomrule
  \end{tabular}
\end{table}

\subsection{The Freezing Phenomenon of Information Flow Structure}
\begin{figure}[h]
\centering
\includegraphics[width=0.6\textwidth]{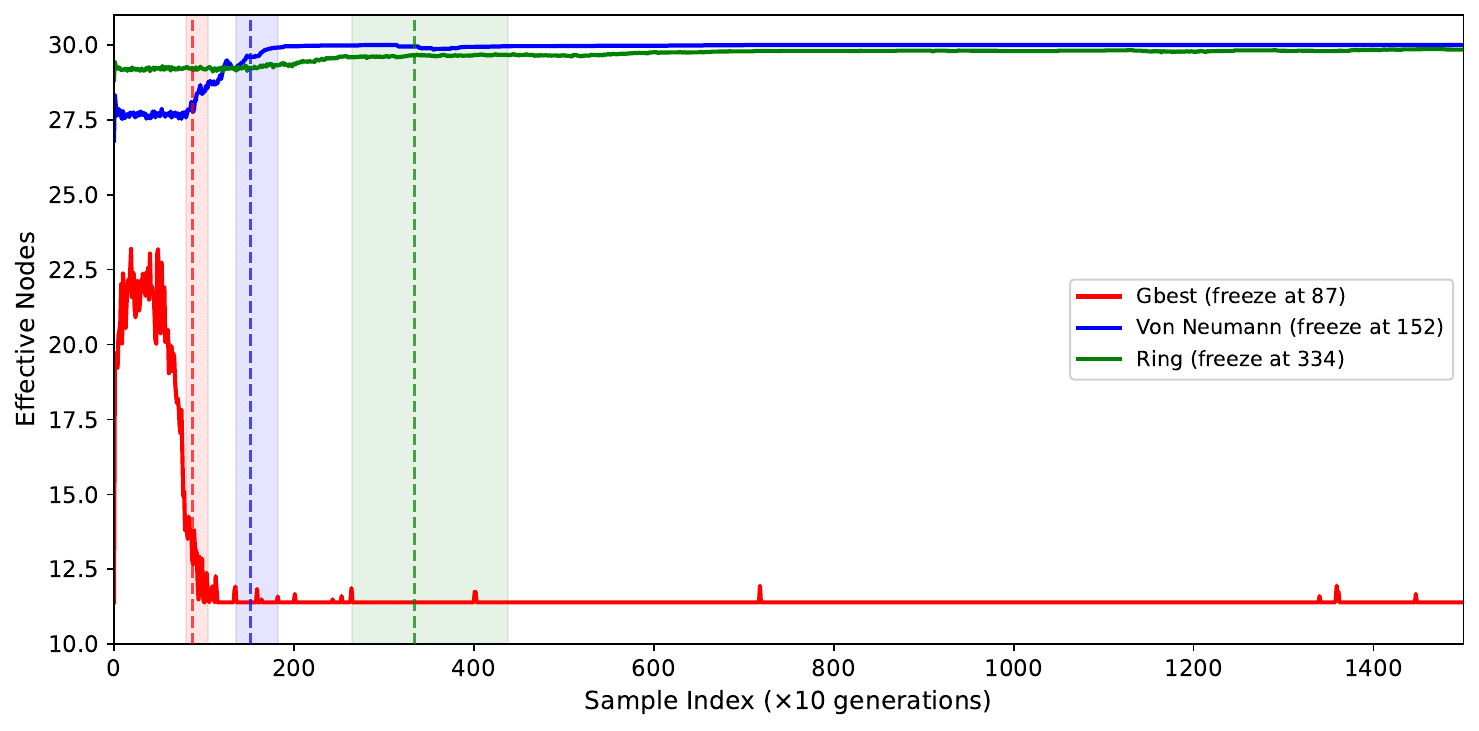} 
\caption{Freezing of Effective Nodes under three topologies for a representative configuration (F8, $30D$, standard PSO). Red: Gbest; Blue: Von Neumann; Green: Ring. Dashed lines mark the median freeze points across 15 runs; shaded bands indicate the 25th–75th percentile range. The global freeze-point distribution across 10 functions and 3 variants is shown in Figure \ref{fig:f4-2}.}
\label{fig:f4-1}
\end{figure}

\begin{figure}[h]
\centering
\includegraphics[width=0.6\textwidth]{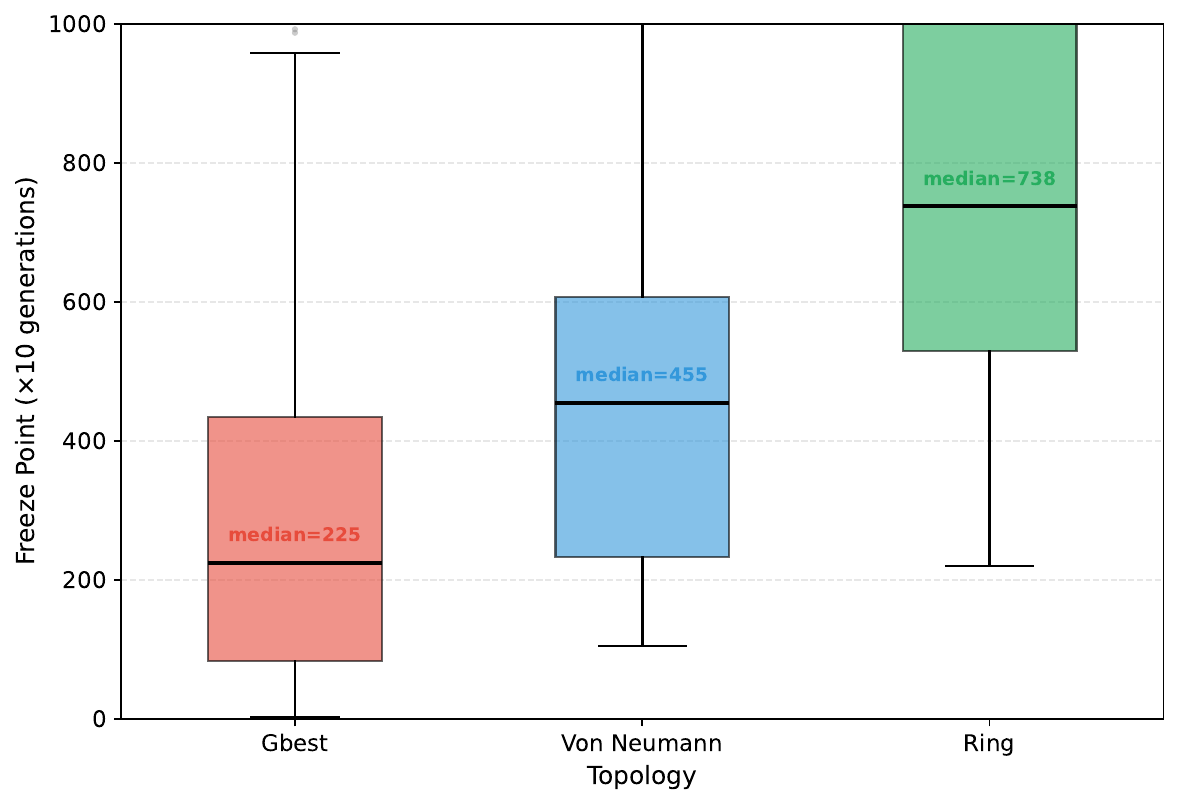} 
\caption{Boxplots of the freeze point distributions under the three topologies ($30D$, based on runs with successful freezing). The detection rates are given in parentheses: Gbest (78.2\%), Von Neumann (36.9\%), and Ring (41.6\%). Runs without detection indicate that the information flow structure had not fully frozen by the end of the search.}
\label{fig:f4-2}
\end{figure}

Figure \ref{fig:f4-1} uses F8 ( $30D$ , standard PSO) as a representative case to show the effective number of nodes freezing curves for the three topologies. Under this representative configuration, the median freeze points for the three topologies are 87 for Gbest, 152 for Von Neumann, and 334 for Ring (sampling points). The global freeze point distributions across all 10 functions and 3 variants are shown in Figure \ref{fig:f4-2}. All three topologies exhibit a clear freezing phenomenon: after an initial period of fluctuation, the Effective Nodes curve becomes completely flat and remains unchanged for the remainder of the search.

As shown in Figure \ref{fig:f4-1}, the Effective Nodes exhibits distinct freezing behaviors across the three topologies. Under the Gbest topology (red), it peaks at approximately 23 during the first 50 samples, then drops rapidly to approximately 11.4 and freezes at about 5\% of the total iterations. Under the Von Neumann topology (blue), it remains near 30 throughout the search and freezes at approximately 9\% of the total iterations. Under the Ring topology (green), it likewise remains near 30 but freezes later, at approximately 22\% of the total iterations.

The freeze point is visually evident: the curve transitions from noisy fluctuations to complete flatness. Once freezing occurs, the effective number of nodes remains unchanged for the remaining 86\% to 95\% of the search. This indicates that the information flow structure, namely the distribution of influence among particles, has become fully fixed.

The freezing phenomenon is not confined to specific functions. Figure \ref{fig:f4-2} shows the freeze-point distribution across all 10 functions and 3 variants. For the representative configuration (F8, $30D$, standard PSO), the median freeze points are Gbest 87, Von Neumann 152, and Ring 334 (as shown in Figure \ref{fig:f4-1}). Aggregated across all configurations, the global median freeze points are Gbest 225, Von Neumann 455, and Ring 738 (as shown in Figure \ref{fig:f4-2}). Although the global values are higher than the representative configuration, the topology order Gbest < Von Neumann < Ring remains consistent in both statistical scopes. 

To verify the robustness of the freezing phenomenon, we also examined the evolution of two other core macro metrics: $G_{s}$ (Strength Gini) and $H_{s}$ (Strength Entropy). All three metrics exhibit consistent freezing behavior: under all three topologies, $G_{s}$ rises to a stable value and then remains constant, while $H_{s}$ decreases to a stable value and then remains constant, with freeze points differing from that of $E_{\text{eff}}$ by less than 10\%. After freezing, the variation of all three metrics is less than 5\%. This indicates that the freezing phenomenon is not an artifact of a single metric but a general manifestation of information flow concentration. Complete evolution curves of the three metrics are provided in Figure S2 of the supplementary material.

\subsection{Topology Dependence of Freeze Time}
Table \ref{tab:4} summarizes the freeze-point statistics for each topology ($30D$, across all 10 functions and 3 variants). Freeze points were detected independently for each run, and the median is reported over all runs in which freezing was successfully detected. The freeze time exhibits a clear order: Gbest freezes earliest, followed by Von Neumann, and Ring freezes latest. This order is consistent with that observed for the representative configuration in Figure \ref{fig:f4-1}, although the absolute values are larger because the statistics here aggregate all functions and variants.

\begin{table}[htbp]
  \centering
  \small
  \setlength{\tabcolsep}{3.5pt} 
  \caption{Statistics of freezing points (FP) grouped by topology ($30D$, across 10 functions and 3 variants)}
  \label{tab:4}
  \begin{tabular}{lccc}
    \toprule
    Topology & Median FP ($\times 10$ gens) & Iteration ratio & Detection rate \\
    \midrule
    Gbest        & 225 & 15.0\% & 78.2\% \\
    Von Neumann  & 455 & 30.3\% & 36.9\% \\
    Ring         & 738 & 49.2\% & 41.6\% \\
    \bottomrule
  \end{tabular}
\end{table}

The median freeze points reported in the Table \ref{tab:4} reveal two noteworthy phenomena. First, the detection rate differs markedly across topologies. Under Gbest, 78.2\% of runs allow a freeze point to be detected, whereas the detection rates under Von Neumann and Ring decrease to 36.9\% and 41.6\%, respectively. Second, the proportion of runs that never froze increases from Gbest to Ring. These two phenomena indicate that the more uniform the topology, the more difficult it is for the information flow structure to become fully frozen. Under Gbest, the star-shaped structure rapidly concentrates information on the global-best particle, yielding a clear and early freeze. Under Ring and Von Neumann, the uniform connectivity pattern suppresses hub dominance, and the information flow structure continues to evolve slowly throughout the search, so that no freeze point is detected before the search ends.

This interpretation is supported by the fact that, among runs with successful freeze detection, the freeze points of Ring and Von Neumann (median 738 and 455 sampling points, respectively) occur much later than that of Gbest (median 225 sampling points).

To verify the cross-function consistency of this topology order, Figure S3 in the Supplementary Material shows the freeze-point distribution box plots for three representative functions (F8, F20, F25). The freeze-point order is Gbest < Von Neumann < Ring in all three functions, further confirming that the topology dependence of freeze time does not change with the function.

\subsubsection{Effect of PSO Variants on Freeze Time}
Freeze time is also influenced by the PSO variant. Table \ref{tab:5} reports the mean freeze point for three variants under each topology on F8 ($30D$).

\begin{table}[htbp]
  \centering
  \setlength{\tabcolsep}{5pt} % 控制列间距，防止表格过宽
  \caption{Median freeze points under different variants and topologies (F8, $30D$; unit: sampling index $\times 10$ generations)}
  \label{tab:5}
  \begin{tabular}{lccc}
    \toprule
    Topology & Standard & LDIW & TVAC \\
    \midrule
    Gbest        & 87  & 156 & 201 \\
    Von Neumann  & 152 & 509 & 284 \\
    Ring         & 334 & 594 & 615 \\
    \bottomrule
  \end{tabular}
\end{table}

As shown in Table \ref{tab:5}, two observations are noteworthy. First, across all three topologies, the LDIW and TVAC variants exhibit later freeze points than standard PSO, indicating that parameter-adaptive strategies delay structural freezing by maintaining population diversity over a longer period. Second, the topological order Gbest < Von Neumann < Ring remains consistent across the three variants. These results indicate that topology determines the relative ordering of freeze times, whereas the variant modulates the overall scale. Given the large variance across independent runs, with coefficients of variation often exceeding 50\%, these values should be interpreted as trend indicators rather than precise estimates.

\subsection{Correlation Between Freeze Point and Search Performance}

\begin{figure}[htbp]
\centering
\includegraphics[width=0.6\textwidth]{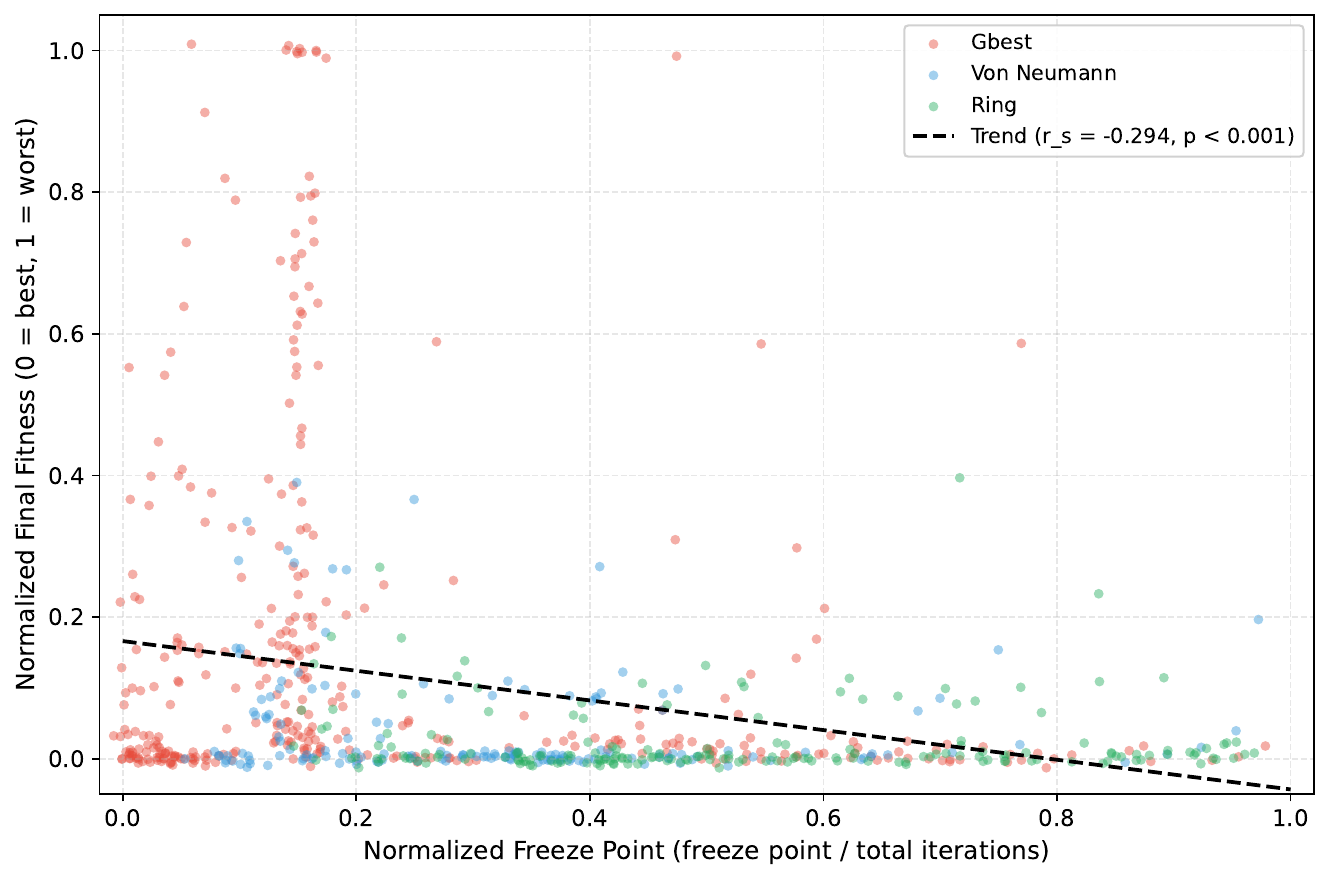} 
\caption{Scatter plot of freeze point versus final fitness ($30D$, all configurations). X-axis: normalized freeze point (freeze point / total iterations); Y-axis: min-max normalized final fitness per function. Each point represents one independent run (705 points in total), colored by topology: red for Gbest, blue for Von Neumann, and green for Ring. The black dashed line shows the linear trend, with Spearman correlation $r_{s} = -0.294$ ($p < 0.001$) . Runs without a detected freeze point are excluded.}
\label{fig:f4-3}
\end{figure}

Figure \ref{fig:f4-3} presents scatter plots of the freeze point versus final solution quality across all configurations. The spearman correlation coefficient is $r_{s}$=$-0.294$, $(p<0.001)$, indicating a significant negative correlation: the earlier the freezing, the worse the final solution quality. This correlation holds for both $10D$ and $30D$ and across all function types. This quantitatively establishes, for the first time, a link between structural freezing and search performance. The correlation is moderate rather than strong, which is understandable: freezing is a structural precursor to performance degradation, but other factors, such as the quality of the frozen configuration, also influence the final outcome.

\subsection{Causal Validation via Intervention Experiments}
The correlation analysis in Section 4.4 establishes an association between freeze point and performance, but cannot determine causality. To establish whether hub dominance causes early freezing and performance degradation, we conducted intervention experiments under the Gbest topology on F8, F15 and F25.

\subsubsection{Intervention Design}
Two opposite interventions were designed:

•   Group A (Hub Weakening): When Effective Nodes drops below 85\% of its peak value, the hub node’s personal best is replaced with a low-fitness particle’s personal best. This weakens the hub’s influence and encourages diversification.

•	Group B (Hub Strengthening): At 5\% of total iterations, 30\% of particles are forced to learn from the hub node. This accelerates information aggregation toward the hub.

•	Control: Standard PSO with no intervention.

All three groups used the same random seeds to ensure fair comparison. Intervention experiments were conducted on F8 (multimodal) and F15 (hybrid), which represent two typical landscape categories. F25 (composition function) was excluded from the main text analysis because it exhibited insensitivity to all interventions, with final fitness differences among the three groups below 0.1 and $p > 0.3\%$.

\subsubsection{Structural Effects}

\begin{figure}[htbp]
\centering
\includegraphics[width=0.9\textwidth]{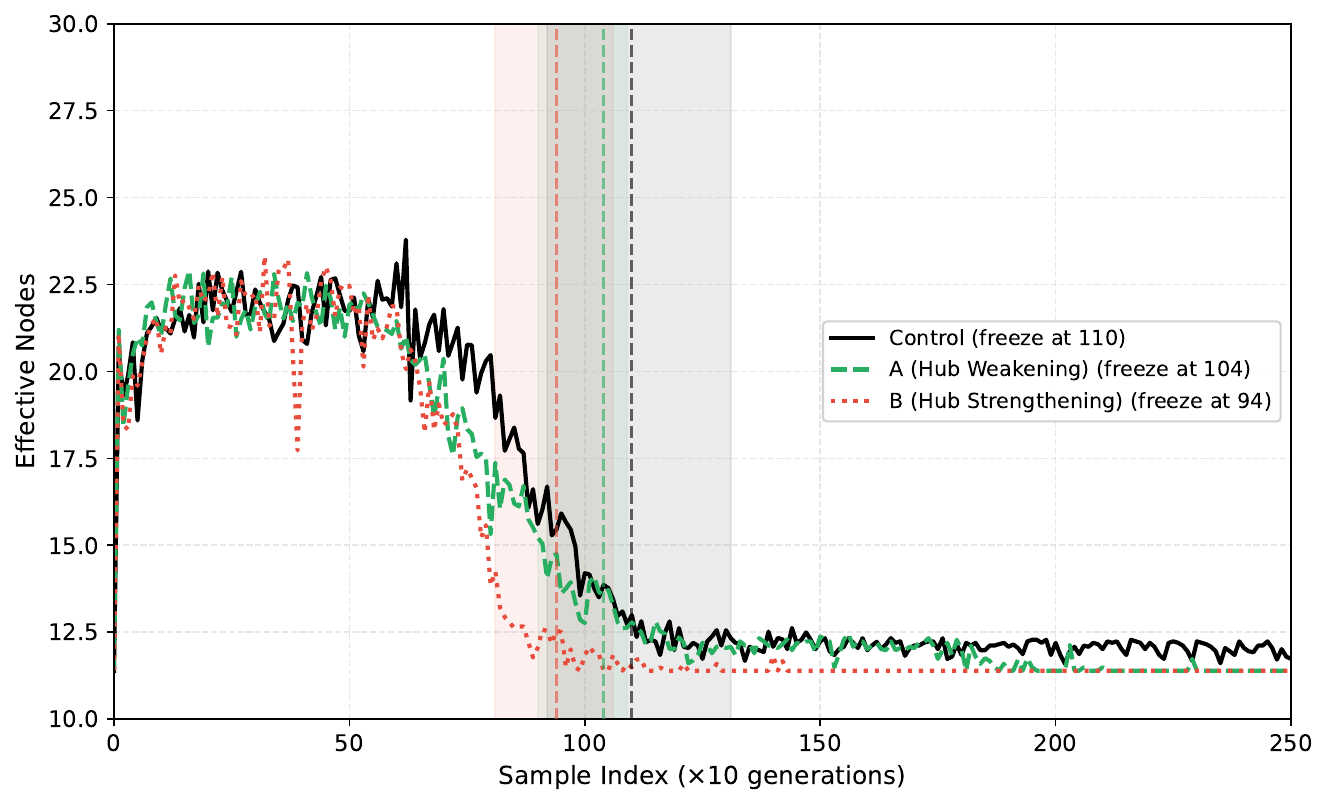} 
\caption{Comparison of Effective Nodes evolution under intervention experiments (F8, $30D$, Gbest). Black: Control; Green: Group A (Hub Weakening); Red: Group B (Hub Strengthening). Dashed lines mark the median freeze points across 15 runs; shaded bands indicate the 25th–75th percentile range. }
\label{fig:f4-4}
\end{figure}

Figure \ref{fig:f4-4} shows the evolution of the effective number of nodes for the three groups, Group B freezes earliest (median 94), while Groups A and Control freeze at 104 and 110, respectively. Group B (hub strengthening) froze earliest (median 94), whereas Group A (hub weakening) and the Control group had freeze points of 104 and 110, respectively. The difference between Group A and the Control group was small (approximately 6 sampling points), while Group B froze significantly earlier than both groups (approximately 16 sampling points). This indicates that under the Gbest topology, hub strengthening significantly accelerates structural freezing, whereas the delaying effect of hub weakening is limited. This is consistent with the saturation effect. Specifically, the star-shaped structure of Gbest is already in an extreme state of hub dominance, and weakening a single hub cannot fundamentally alter the overall structural trajectory. Supplementary Table S2 shows that the group-wise structural differences in the three core macro metrics ($E_{\text{eff}}$, $G_{s}$, $H_{s}$) are small. Specifically, these differences are completely identical on F8, with zero standard deviation, and remain within one standard deviation on F15. This is consistent with the small freeze-point differences in Figure \ref{fig:f4-4}, jointly indicating that under the Gbest saturated topology, the structural effect of intervention is very limited.

Notably, although the structural effect of Group A is weak, its final fitness is significantly better than that of the Control group (Table 4, $p=0.0009$). This indicates that under a saturated topology, the benefits of intervention are achieved primarily by altering the learning behavior of particles rather than by changing the information flow structure. In other words, structural effects and performance effects can be decoupled, such that in a structurally saturated system, fine-tuning at the behavioral level can still yield significant performance improvements.

\subsubsection{Performance Effects}

\begin{table}[htbp]
  \centering
  \small
  \setlength{\tabcolsep}{5pt}
  \begin{threeparttable}
    \caption{Intervention experiment results}
    \label{tab:6}
    \begin{tabular}{l c c c}
      \toprule
      Function & Group & Final fitness (mean$\pm$std) & vs Control (p-value) \\
      \midrule
      F8 & Control & $813.23 \pm 4.45$ & --- \\
      F8 & A (weaken) & $807.88 \pm 4.24$ & $0.0009\downarrow$ \\
      F8 & B (strengthen) & $808.35 \pm 3.34$ & $0.9973$ \\
      F15 & Control & $1256.7 \pm 466.7$ & --- \\
      F15 & A (weaken) & $1092.8 \pm 466.2$ & $0.18$ \\
      F15 & B (strengthen) & $1133.3 \pm 418.0$ & $0.71$ \\
      \bottomrule
    \end{tabular}
    \begin{tablenotes}
      \small
      \item All $p$-values are based on one-tailed Mann-Whitney U test, with 15 independent runs per group. $\downarrow$ indicates that Group A is significantly better than the control group.
    \end{tablenotes}
  \end{threeparttable}
\end{table}

Table \ref{tab:6} reports the final fitness of the three groups on F8 and F15. As shown in the table, on F8, the final fitness of the hub-weakening group (Group A) is significantly better than that of the Control group (807.88 vs 813.23, $p=0.0009$), confirming the inhibitory effect of hub dominance on multimodal optimization. The final fitness of the hub-strengthening group (Group B) does not differ significantly from that of the Control group ($p=0.9973$), showing no additional performance deterioration. This asymmetric result is consistent with the saturation effect: the Gbest topology is already in an extreme state of hub dominance, and further strengthening hubs cannot produce additional negative effects.

On F15, the final fitness values of Groups A and B are lower than that of the Control group (A: 1092.8, B: 1133.3, Control: 1256.7), indicating that both interventions slightly improve performance. The improvement is larger in Group A, but neither difference reaches statistical significance (A vs. Control: $p=0.18$; B vs. Control: $p=0.71$). This result is not entirely consistent with that on F8, where Group A improved significantly while Group B showed no change, indicating that the intervention effect is function-dependent. This may be because the hybrid function landscape is relatively smooth and less sensitive to hub dominance than multimodal functions, resulting in a weaker intervention effect. It also suggests that the intervention effect depends on the complexity of the fitness landscape.

It is worth noting that in Figure \ref{fig:f4-4} (structural effect), the difference in freeze points between Group A and the Control group is very small (104 vs. 110), whereas the performance improvement of Group A in Table \ref{tab:6} is highly significant. This contrast indicates that under a saturated topology, the benefits of intervention are achieved mainly by altering the learning behavior of particles rather than by changing the information flow structure. In other words, structural effects and performance effects can be decoupled: even if the information flow structure is insensitive to intervention, fine-tuning at the behavioral level can still yield significant performance improvements.

Supplementary Table S2 further compares the group differences in $G_{s}$ and $H_{s}$. On F8, the three groups exhibit identical $G_{s}$ and $H_{s}$ at the end of the search, indicating that the intervention produces no significant structural impact on the information flow distribution. On F15, Group A shows a slightly lower $G_{s}$ ($0.2023$ $vs$ $0.2048$) and a slightly higher $H_{s}$ ($3.2782$ $vs$ $3.2618$) than the Control, but the differences are much smaller than the performance differences. These results are consistent with the small freeze-point differences in Figure \ref{fig:f4-4}, jointly indicating that under the Gbest saturated topology, the structural impact of intervention on the information flow distribution is very limited. The benefit of intervention is primarily achieved by altering the learning behavior of particles rather than by changing the information flow structure. 

\subsubsection{The Saturation Effect}
The asymmetry between Groups A and B, together with the contrast between Figure \ref{fig:f4-4} and Table \ref{tab:6}, jointly reveals a more complete mechanism. Under the Gbest topology, all particles already learn from the global best particle, and the information flow structure is maximally centralized ($E_{\text{eff}}$ $\approx 11.4$, far below $N=30$). This saturated state has two consequences. 

First, the structural channel is closed. Hub strengthening cannot further increase information flow concentration (in Figure \ref{fig:f4-4}, Group B differs from the Control by only 16 sample points), and hub weakening cannot significantly reverse the structural concentration trend (Group A differs from the Control by only 6 sample points). Supplementary Table S2 shows that the group differences in the three core macro metrics ($E_{\text{eff}}$, $G_{s}$,$H_{s}$) are close to zero on F8, further confirming that the room for structural intervention is extremely limited.

Second, the behavioral channel remains effective. Despite the weak structural effect, the hub-weakening group achieves significantly better final fitness than the Control (Table \ref{tab:6}, $p = 0.0009$). This indicates that under a saturated topology, the benefit of intervention is primarily achieved by altering the learning behavior of particles. Once the hub's pbest is replaced, particles obtain new learning exemplars and escape the original information aggregation path.

This phenomenon is termed the saturation effect: when the topology itself already operates at the extreme of hub dominance, further strengthening hubs produces no additional effect, and weakening hubs cannot easily change the overall structure, yet behavioral-level adjustments can still yield significant performance improvements. The saturation effect reveals a decoupling between structural effects and performance effects. Specifically, in a structurally saturated system, performance improvement does not require structural adjustment, and behavioral intervention alone suffices.

This finding explains why Gbest is inherently prone to premature convergence: it is structurally locked into a configuration that maximizes short-term convergence at the cost of long-term exploration. It also offers new insights for adaptive topology design: under saturated topologies, rather than attempting to adjust the structure, it is more effective to intervene at the behavioral level (e.g., replacing learning exemplars or introducing diversity perturbations).

\subsection{Micro-Level and Meso-Level Auxiliary Analysis}
The main analyses in Sections 4.2 to Section 4.5 are based on the macro-level metric, namely the effective number of nodes. Here, we report supplementary results for the micro-level and meso-level metrics, which provide additional perspectives on the freezing mechanism.

\subsubsection{Micro-Level Analysis: Hub Formation}

\begin{figure}[htbp]
\centering
\includegraphics[width=0.7\textwidth]{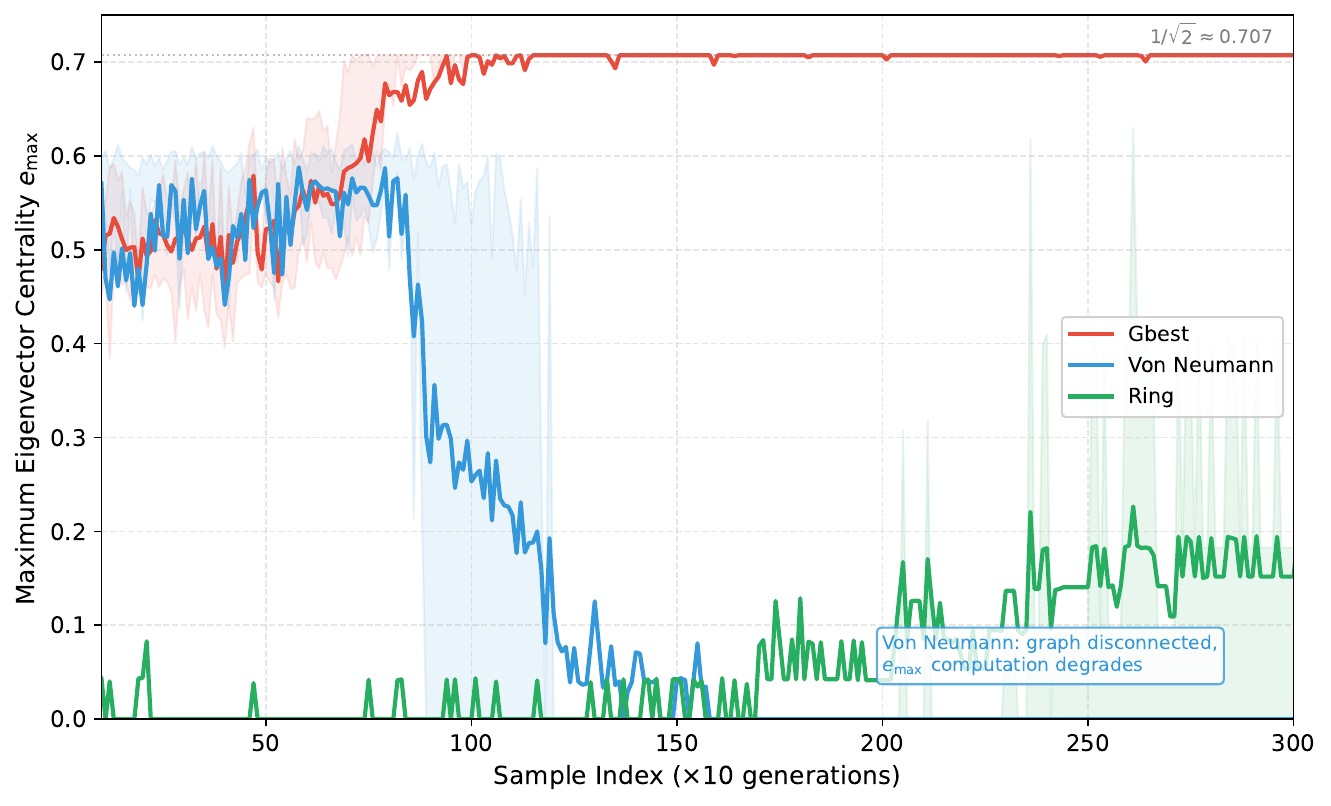} 
\caption{Evolution of maximum eigenvector centrality $e_{\max}$ under three topologies (F8, $30D$, standard PSO). Red: Gbest; Blue: Von Neumann; Green: Ring. Shaded bands indicate the 25th–75th percentile range across 15 runs.}
\label{fig:f4-5}
\end{figure}

Figure \ref{fig:f4-5} shows the evolution of maximum eigenvector centrality $e_{\max}$ under three topologies. All three topologies exhibit sharp structural transitions around sample 90, synchronized with the Gbest freeze point (87) in Figure \ref{fig:f4-1}.

Under Gbest, $e_{\max}$ rises sharply to 0.7071 (the theoretical value for a star graph) around sample 90, indicating that hub dominance is fully established at this point. This jump is synchronized with the Gbest freeze point, confirming the mechanism of “structural freezing corresponds to hub saturation.”

Under Ring, $e_{\max}$ drops to 0 around sample 90 and then slowly fluctuates in the 0.15–0.20 range. This is consistent with the uniform connection pattern of Ring: each particle connects only to two neighbors, centrality is uniformly distributed across nodes, and no dominant hub can form.

Under the Von Neumann topology, $e_{\max}$ falls to zero near sample 90. This degradation arises from the decreased connectivity of the influence graph in the later stages of the search, which renders eigenvector-based centrality computation unreliable. The micro-level analysis of this topology is therefore not pursued in this paper.

The sharp transitions near sample 90 are synchronized with the Gbest freeze point, further supporting the conclusion that structural freezing is intrinsically linked to hub evolution.

\subsubsection{Meso-Level Analysis: Community Structure}

\begin{figure}[htbp]
\centering
\includegraphics[width=0.7\textwidth]{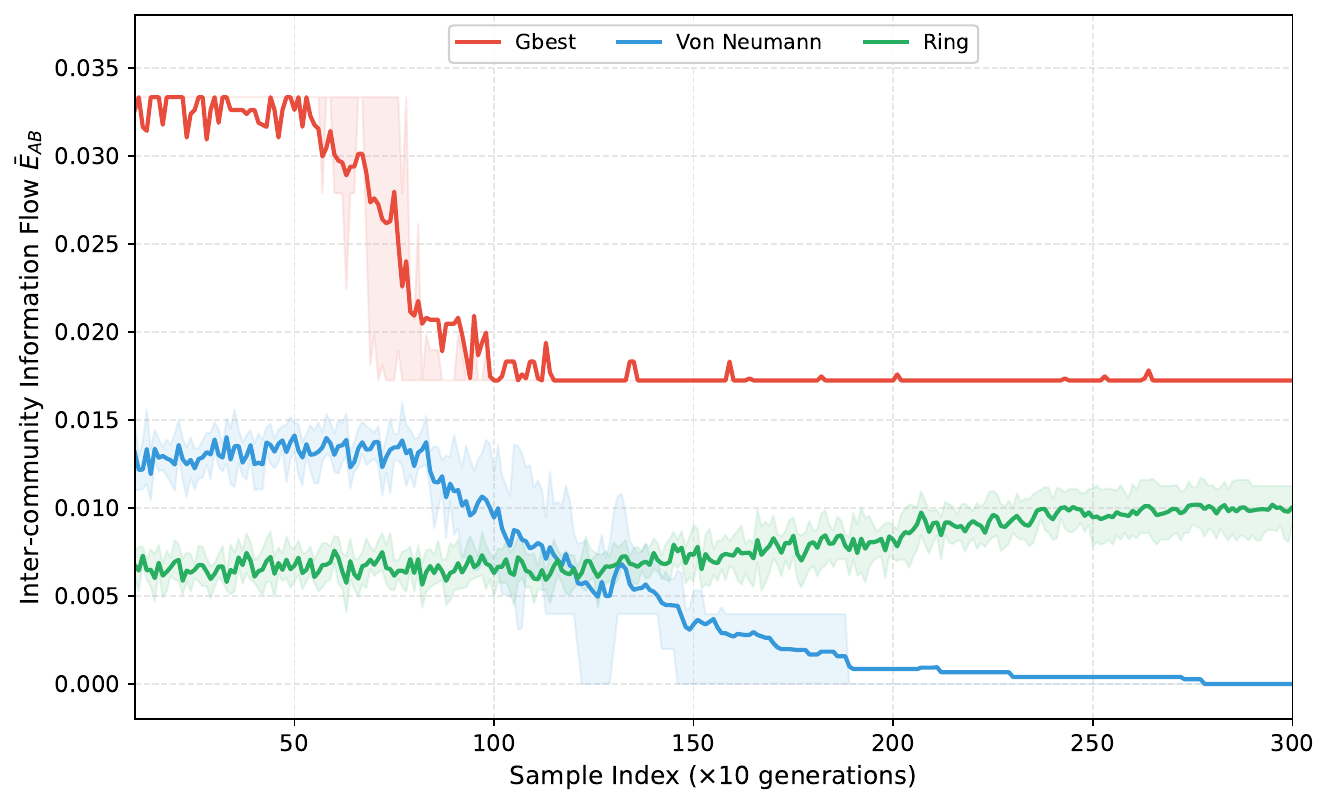} 
\caption{Evolution of inter-community information flow $\bar{E}_{AB}$ under three topologies (F8, $30D$, standard PSO). Red: Gbest; Blue: Von Neumann; Green: Ring. Shaded bands indicate the 25th–75th percentile range across 15 runs.}
\label{fig:f4-6}
\end{figure}

Figure \ref{fig:f4-6} shows the evolution of inter-community information flow $\bar{E}_{AB}$ under three topologies. The magnitude of $\bar{E}_{AB}$ remains below 0.03 across all three topologies, far smaller than the value ranges of macro-level metrics ($E_{\text{eff}}$ $\approx 11 - 30$) and micro-level metrics ($e_{\max}$ $\approx 0-0.7$). We note that this finding is consistent with Section 3.3.2, where we showed that under regular topologies, community structure is largely predetermined by the topology itself and does not evolve substantially during the search.

Under Gbest, $\bar{E}_{AB}$ drops sharply from 0.033 to 0.017 between samples 80 and 90, and then remains stable. This time point is again synchronized with the Gbest freeze point (87). This indicates that the meso-level metric also undergoes a structural transition at the freeze point, which is consistent with the synchronized jump of the micro-level metric $e_{\max}$  shown in Figure \ref{fig:f4-5}. After the freeze point, the star graph has no community structure, and $\bar{E}_{AB}$ stabilizes at a low value.

Under Ring, $\bar{E}_{AB}$ slowly rises from approximately 0.007 to 0.012. We note that this behavior is consistent with the uniform connection pattern of the Ring topology. Under this topology, the local community structure evolves slowly during the search, and cross-community exchange increases only modestly, with the overall magnitude remaining limited. Under Von Neumann, $\bar{E}_{AB}$  gradually declines from approximately 0.013 to 0, indicating that community structure becomes fully isolated in the later search stage.

The weak variation of the meso-level metric supports the metric screening logic adopted in this paper. Under regular topologies, meso-level metrics are not used as the primary basis for freeze-point detection but serve only as auxiliary verification. Notably, the sharp drop of the meso-level metric under Gbest is synchronized with the freeze point, further confirming the qualitative finding that the freeze point represents a key moment of synchronized structural transitions across multiple scales.

\subsection{Summary}
This section reveals three core findings.

(1)	PSO information flow structure freezes at an early stage, after which it remains completely unchanged. This freezing phenomenon is observed across all 10 CEC2017 functions and 3 PSO variants.

(2)	Freeze time is determined by topology (Gbest < Von Neumann < Ring), while parameter-adaptive variants (LDIW, TVAC) delay freezing without altering the topology order.

(3)	Freeze point is significantly negatively correlated with performance. Furthermore, intervention experiments confirm the causal chain, showing that weakening hubs delays freezing and improves performance, whereas strengthening hubs produces no additional effect due to saturation.

We also find that the micro-level and meso-level auxiliary analysis provides complementary evidence, in that Gbest exhibits saturated hub dominance from the start, whereas Ring and Von Neumann exhibit gradual hub formation. Meso-level metrics confirm that community structure is largely predetermined by regular topologies and does not evolve substantially.

\section{Discussion}
\subsection{Mechanistic Interpretation of Structural Freezing}
Why does the information flow structure freeze? Under the Gbest topology, all particles learn from the global best. Once the global best is determined or stabilized, the influence graph becomes a fixed star-shaped structure centered on that particle, and it cannot change because no other learning sources are considered. Under the Ring and Von Neumann topologies, particles learn from local neighbors. Once the best neighbor of each particle is determined, the edge set is likewise fixed. In all cases, structural freezing reflects the convergence of learning relationships among particles.

This freezing phenomenon explains a long-standing puzzle in PSO research, namely why the algorithm appears to still be running, with fitness continuing to improve slowly, when it has in fact ceased exploring. Structural freezing occurs at an early stage (3\% to 22\% of the total iterations under the representative configuration and at 15\% to 49\% across all configurations), implying that 51\% to 85\% of the computational budget is wasted on redundant iterations over a frozen network. This is a fundamental structural inefficiency.

Notably, freezing is not an event at a single scale but a synchronized multi-scale transition. Both Figure \ref{fig:f4-5} (micro-level $e_{\max}$) and Figure \ref{fig:f4-6} (meso-level $\bar{E}_{AB}$) exhibit sharp transitions around samples 80 to 90, synchronized with the Gbest freeze point (87) in Figure \ref{fig:f4-1} (macro-level $E_{\text{eff}}$). We observe that the core metrics at the three scales undergo structural transitions simultaneously near the freeze point, with the macro-level metric ceasing to decrease, the micro-level metric jumping to a saturated value, and the meso-level metric dropping sharply. This indicates that the freeze point is not only the moment at which the macro-level distribution solidifies but also a critical point at which the entire population structure, including nodes, communities, and the global scale, is synchronously adjusted. This cross-scale synchronization provides multi-perspective evidence that the freeze point is a structural phase transition.

\subsection{Topology Dependence and Its Implications}
The freeze time order Gbest < Von Neumann < Ring directly reflects the information propagation speed of each topology. Gbest has the shortest path length (all particles connect to one center), so information propagates fastest and the structure freezes earliest. Ring has the longest path length (each particle connects only to two neighbors), so information propagates slowly and freezing is delayed. Von Neumann, with four neighbors per particle, lies in between.

This provides a structural explanation for the classic observation that Gbest performs well on unimodal functions but poorly on multimodal functions. On unimodal functions, rapid freezing is beneficial because there is only one optimum to find. On multimodal functions, early freezing is harmful because the population loses the ability to explore alternative regions. We therefore interpret the freeze time as a structural indicator of the exploration–exploitation trade-off, in that topologies with early freezing are biased toward exploitation, whereas those with late freezing maintain exploration capability for longer.

Freeze time is also influenced by the PSO variant. Under all three topologies, the freeze points of LDIW and TVAC are generally later than that of standard PSO, indicating that parameter-adaptive strategies maintain population diversity for a longer period by dynamically adjusting the inertia weight or acceleration coefficients, thereby delaying structural freezing. Under the Gbest, Von Neumann, and Ring topologies, both LDIW and TVAC significantly delay freezing; for example, under Ring topology, the freeze point is delayed from 334 to 580–613. This is consistent with the expectation that parameter-adaptive strategies delay structural freezing by maintaining diversity.

\subsection{Causal Evidence and the Saturation Effect}
Intervention experiments provide the strongest evidence for the causal chain. Weakening hubs significantly improves performance ($p=0.0009$ on F8), whereas strengthening hubs produces no additional deterioration. This asymmetry reveals the saturation effect. Specifically, the Gbest topology already operates at the maximum level of hub dominance, and the information flow structure cannot become further centralized.

More significantly, the comparison between Figure \ref{fig:f4-4} and Table \ref{tab:6} reveals a decoupling between structural effects and performance effects. Under the Gbest topology, Group A and the Control group differ by only 6 sampling points in their freeze points, and the between-group differences in the three core macro-level metrics  ($E_{\text{eff}}$, $G_{s}$, $H_{s}$) on F8 are close to zero. However, the final fitness of Group A is significantly better than that of the Control group ($p=0.0009$). This indicates that under a saturated topology, the benefit of intervention is achieved primarily by altering the learning behavior of particles rather than by changing the information flow structure.

This finding has direct implications for adaptive topology design. Under saturated topologies, the structural channel is effectively closed, and structural adjustment therefore yields only limited gains. It is thus preferable to improve performance through behavioral-level interventions, such as replacing learning exemplars and introducing diversity perturbations. The saturation effect reveals the boundary conditions of structural intervention. Specifically, when the topology has already reached an extreme state of hub dominance, the behavioral channel is more effective than the structural channel.

\subsection{Why Only Macro-Level Metrics Capture the Freezing Phenomenon}
A notable finding of this study is that, among the metrics at the three levels, only the macro-level metric can effectively characterize the freezing phenomenon. Micro-level metrics, such as maximum eigenvector centrality, and meso-level metrics, such as inter-subgroup information flow, provide useful auxiliary information, but they cannot serve as the primary basis for freeze-point detection. This limitation is not a deficiency in metric design but an inherent consequence of the structural properties of PSO populations under regular topologies.

First, under regular topologies, the micro-level and meso-level structures are predetermined by the topology itself and do not evolve during the search. Under the Gbest, Ring, and Von Neumann topologies, node degrees are fixed, and all particles possess the same number of neighbors, with the exception of the central node in Gbest. Community structure is likewise predetermined. The star graph contains no communities, the ring graph exhibits no significant communities, and the grid graph has a fixed community structure. Consequently, the micro- and meso-level metrics remain essentially unchanged throughout the search process.

Second, the information capacity of small-scale networks is limited. When $N=30$, node-level metrics provide only 30 observations, and these are highly correlated. The resolution limit of modularity, approximately 6.7 edges when $N=30$ and the average degree is 6, renders community detection unstable. In contrast, macro-level metrics aggregate all edge weights and extract discriminative information from the entire distribution rather than from individual nodes or communities.

Third, hub dominance is inherently a macro-level distributional phenomenon. It is not that the degree of a particular node becomes large, which is a micro-level property, but that information flow becomes concentrated on a small number of nodes, which is a distributional property. This can only be quantified by distributional metrics such as the effective number of nodes, the Gini coefficient, and entropy.

Fourth, the topological differences in freeze detection rates also support this conclusion. Under Gbest, 78.2\% of runs allow a freeze point to be detected, whereas the detection rates under Von Neumann and Ring decrease to 36.9\% and 41.6\%, respectively. This indicates that the information flow structure is more difficult to fully freeze under uniform topologies, which is consistent with the mechanism that micro-level and meso-level structures are predetermined by the topology and that uniform topologies lack a driving force for centralization.

We believe this finding carries methodological implications. Specifically, we suggest that macro-level distributional metrics are more discriminative than micro-level and meso-level structural metrics when analyzing dynamic networks under regular topologies. For future work, if micro-level and meso-level metrics are to be used, we recommend considering non-regular topologies such as random graphs and scale-free networks, or larger population sizes, so that the micro-level and meso-level structures have sufficient degrees of freedom to evolve.

\subsection{Comparison with Existing Work}
In contrast to the influence graph framework and diagnostic metrics provided by Oliveira et al. \cite{oliveira2013assessing,oliveira2015using,oliveira2016communication,oliveira2014towards}, the present work moves beyond a posteriori diagnosis toward mechanism revelation. Their work established how to model a PSO population as a network and how to employ network metrics for stagnation detection. Building on this foundation, this paper addresses a different question: not whether the population has stagnated, but when and why the information flow structure ceases to change.

Compared with Deng et al. \cite{deng2025collective}, who described the static topological properties of PSO populations, this work characterizes dynamic structural evolution and identifies the freeze point as a key transition. The small-world architecture and heavy-tailed degree distribution reported by Deng et al. \cite{deng2025collective} provide a static snapshot, whereas the early freezing revealed in this work provides the dynamic trajectory that gives rise to that snapshot.

In contrast to adaptive topology design approaches such as SpadePSO, Bifurcated PSO, RCI, and HPSO, this work does not introduce new topology rules. Rather, it uncovers the structural evolution patterns of existing regular topologies. A common feature of these approaches is that they directly target algorithmic performance improvement by incorporating new learning indicators, particle roles, or neighborhood rules to enhance search behavior. They do not analyze the structural evolution of the topology itself, nor do they address the fundamental question of when the information flow structure ceases to change. The findings of this work provide a mechanistic foundation for such adaptive topology designs, in which the freeze point can serve as a triggering signal for topology adjustment.

Compared with Gupta et al. \cite{gupta2025enhancing}, who used an explainability framework to systematically compare how Ring, Star, and Von Neumann topologies affect PSO convergence and search behavior, revealing how these topologies influence information flow, diversity, and convergence speed. This work is closest to ours in terms of research object, as both focus on these three topologies and address the relationship between information flow and search behavior. However, the analytical levels differ fundamentally. Gupta et al. provide a static comparison of topology types, answering how each topology performs, whereas this paper reveals the dynamic evolution mechanism of information flow structure within a given topology, answering when the structure freezes, how freeze time depends on topology, and whether there is a causal link between freezing and performance. Furthermore, Gupta et al. focus on building an explainability framework, whereas this paper focuses on mechanism discovery and causal validation, establishing the causal chain from topology to freezing and then to performance through intervention experiments.

The novel contribution of this work lies in the discovery and systematic characterization of the freezing phenomenon. Although prior studies have noted that PSO populations exhibit small-world and heavy-tailed properties, the observation that these structures freeze early and that the freeze time determines performance has not yet been reported.

\subsection{Scope and Generalizability}
This study focuses on small-scale PSO with 20 to 50 particles, regular topologies (Gbest, Ring, and Von Neumann), and conventional variants (standard, LDIW, and TVAC). This scope is selected for three reasons. First, a population size of 20 to 50 particles is the most common in practical PSO implementations. Second, regular topologies provide controlled experimental conditions that facilitate isolating the influence of topology on structural evolution. Third, conventional variants enable us to examine the freezing phenomenon without interference from complex adaptive mechanisms.

The proposed multi-scale information-flow evaluation framework and freeze-point detection method are topology-independent and can be extended to non-regular topologies and larger population sizes. In large-scale irregular networks, micro-level and meso-level metrics may regain their discriminative power, because node degrees become heterogeneous and community structures evolve dynamically. Under such conditions, the freeze point may be delayed or exhibit multi-stage characteristics. These issues constitute promising directions for future work.

\subsection{Limitations}
This study has several limitations. First, while the experiments are conducted on CEC2017 benchmark functions, validation on real-world optimization problems remains necessary. Second, although the freeze-point detection parameters ($W=50$, $\varepsilon ={10}^{-3}$) were selected based on preliminary experiments, , a more systematic sensitivity analysis would strengthen the method. Third, the intervention experiments were limited to the Gbest topology, and extending them to the Von Neumann and Ring topologies would provide a more complete picture. Fourth, the Von Neumann topology exhibited graph disconnection in the later search stage, which degraded eigenvector-based micro-level analysis, and alternative micro-level metrics for disconnected graphs warrant further investigation.

\section{Conclusion}
\subsection{Main Findings}
This paper investigates the freezing mechanism and topology dependence of information flow structure in small-scale PSO with regular neighborhood topologies. Through systematic experiments on 10 CEC2017 benchmark functions with three topologies (Gbest, Ring, Von Neumann) and three PSO variants (standard, LDIW, TVAC), the following conclusions are drawn.
	
(1) PSO information flow structure freezes at an early stage of the search. The Effective Nodes metric stabilizes within 3\%–22\% of total iterations for representative configurations; 15\%–49\% of total iterations for global medians and remains completely unchanged for the remainder of the search. This freezing phenomenon is observed across all tested functions and variants, indicating that it is an intrinsic structural property of PSO rather than an artifact of specific problems. The sharp transitions of the micro-level metric $e_{\max}$ and the meso-level metric ${\bar{E}}_{AB}$ are synchronized with the macro-level freeze point, indicating that freezing is a key moment of multi-scale synchronized structural transition.

(2) Freeze time is determined by the neighborhood topology and modulated by the algorithm variant. The freeze point follows the order Gbest < Von Neumann < Ring, with mean values of approximately 15\%, 30.3\%, 49.2\% of total iterations, respectively. Parameter-adaptive variants (LDIW, TVAC) delay freezing under all three topologies.

(3)	The freeze point is significantly negatively correlated with search performance. The earlier the freeze, the worse the final solution quality  (Spearman $r_{s}=-0.294$, $p<0.001$). This correlation holds across dimensions and function types, establishing a quantitative link between structural freezing and performance degradation for the first time.

(4) Intervention experiments confirm the causal chain and reveal the saturation effect and structure–performance decoupling. Under the Gbest topology, hub weakening significantly improves performance on multimodal functions ($p = 0.0009$ on F8), while hub strengthening produces no additional deterioration. More notably, Group A differs from the Control by only 6 sample points in freeze time, and the group differences in the three core macro metrics ($E_{\text{eff}}$, $G_{s}$, $H_{s}$) are close to zero, yet the performance improvement is highly significant. This indicates that, under a saturated topology, the benefit of intervention is achieved primarily through the behavioral channel rather than the structural channel. The observed decoupling between structural effects and performance effects further supports the saturation effect.

\subsection{Methodological Implications}
A notable methodological finding of this study is that, among the three levels of metrics, only macro-level distributional metrics effectively capture the freezing phenomenon under regular topologies. Micro-level and meso-level metrics provide auxiliary information but do not serve as the primary basis for freeze detection. This is not a flaw in metric design but a consequence of the structural characteristics of small-scale regular networks. Specifically, node degrees and community structures are predetermined by the topology and do not evolve during the search. In contrast, hub dominance is inherently a distribution-level phenomenon, which can only be quantified by macro-level metrics such as Effective Nodes, the Gini coefficient, and Entropy.

This finding suggests that when analyzing dynamic networks under regular topologies, macro-level distributional metrics are more discriminative than micro-level and meso-level structural metrics. Future work aiming to leverage micro-level and meso-level metrics should consider non-regular topologies or larger population sizes, where these structures have sufficient degrees of freedom to evolve.

\subsection{Limitations and Future Work}
This study has several limitations. First, the scope is restricted to small-scale PSO (20–50 particles) with regular topologies and conventional variants. While this regime covers the most common PSO implementation, the generalizability of the freezing mechanism to large-scale populations and complex network topologies remains to be verified. Second, although the freeze-point detection parameters were selected based on preliminary experiments, a more systematic sensitivity analysis would further strengthen the method. Third, although the intervention experiments were conducted only under the Gbest topology, extending them to the Von Neumann and Ring topologies would provide a more complete picture. Fourth, the Von Neumann topology exhibited graph disconnection in the later search stage, which degraded eigenvector-based micro-level analysis.

Future work can proceed in four directions: (1) developing adaptive topology strategies that trigger interventions at the freeze point rather than at fixed iterations; (2) extending the freeze-point framework to dynamic optimization and multi-swarm PSO; (3) establishing a theoretical model linking topology parameters to freeze time; (4) applying the framework to large-scale populations with complex network topologies, where micro-level and meso-level metrics may regain discriminative power.

%% Use a table environment to create tables.
%% Refer following link for more details.
%% https://en.wikibooks.org/wiki/LaTeX/Tables
%
%% Use \caption command for table caption and label.
%

\section*{Acknowledgements}
This work was supported by the Natural Science Foundation of the Jiangsu Higher Education Institutions of China (Grant No. 22KJB520023), the Foundation of State Key Lab. for Novel Software Technology (Grant No. KFKT2022B35), the Starup Foundation for Talents of Nanjing University of Information Science and Technology, the National Natural Science Foundation of China (Grant No. 62376127, 62106092), and the Natural Science Foundation of Fujian Province (Grant No. 2022J01916). The authors would like to thank the anonymous reviewers  for their valuable comments and suggestions.

% ========== （CRediT） ==========
\section*{CRediT authorship contribution statement}
Zhenglong Xiang: Conceptualization, Methodology, Writing-original draft, Funding acquisition. Hongrun Wu: Methodology, Investigation, Funding acquisition. Yu Xue: Validation, Funding acquisition. Fei Yu: Validation. Yuanxiang Li: Conceptualization.

% ==========  ==========
\section*{Declaration of competing interest}
The authors declare that they have no known competing financial interests or personal relationships that could have appeared to influence the work  reported in this paper.

%% Use figure environment to create figures
%% Refer following link for more details.
%% https://en.wikibooks.org/wiki/LaTeX/Floats,_Figures_and_Captions
%\begin{figure}[h]%% placement specifier
%% Use \includegraphics command to insert graphic files. Place graphics files in 
%% working directory.
%

%% The Appendices part is started with the command \appendix;
%% appendix sections are then done as normal sections
%
%% For citations use: 
%%       \citet{<label>} ==> Lamport (1994)
%%       \citep{<label>} ==> (Lamport, 1994)
%%
%Example citation, See \citet{lamport94}.

%% If you have bib database file and want bibtex to generate the
%% bibitems, please use
%%
%%  \bibliographystyle{elsarticle-harv} 
%%  \bibliography{<your bibdatabase>}

%% else use the following coding to input the bibitems directly in the
%% TeX file.

%% Refer following link for more details about bibliography and citations.
%% https://en.wikibooks.org/wiki/LaTeX/Bibliography_Management

\bibliographystyle{elsarticle-num}
\bibliography{ref}

@inproceedings{kennedy1995particle,
  title={Particle swarm optimization},
  author={Kennedy, James and Eberhart, Russell},
  booktitle={Proceedings of ICNN'95-international conference on neural networks},
  volume={4},
  pages={1942--1948},
  year={1995},
  organization={IEEE}
}

@article{sharma2025survey,
  title={A survey on Particle Swarm Optimization: Evolution, adaptations and practical implementations},
  author={Sharma, Rashmi and Matharu, Jagjit Singh and Parmar, Kulwinder Singh},
  journal={Applied Soft Computing},
  pages={114016},
  year={2025},
  publisher={Elsevier}
}

@article{gad2022particle,
  title={Particle swarm optimization algorithm and its applications: A systematic review},
  author={Gad, Ahmed G},
  journal={Archives of Computational Methods in Engineering},
  volume={29},
  number={5},
  pages={2531--2561},
  year={2022},
}

@article{nigatu2023review,
  author  = {Nigatu, Dereje Tarekegn and Tilahun, Surafel Luleseged and Dinka, Tekle Gemechu},
  title   = {A Review on Convergence Analysis of Particle Swarm Optimization},
  journal = {International Journal of Swarm Intelligence Research},
  year    = {2023},
  volume  = {14},
  number  = {1},
  pages   = {1--34},
}

@article{blackwell2018impact,
  title={Impact of communication topology in particle swarm optimization},
  author={Blackwell, Tim and Kennedy, James},
  journal={IEEE Transactions on Evolutionary Computation},
  volume={23},
  number={4},
  pages={689--702},
  year={2018},
  publisher={IEEE}
}

@inproceedings{kennedy2002population,
  title={Population structure and particle swarm performance},
  author={Kennedy, James and Mendes, Rui},
  booktitle={Proceedings of the 2002 Congress on Evolutionary Computation. CEC'02 (Cat. No. 02TH8600)},
  volume={2},
  pages={1671--1676},
  year={2002},
  organization={IEEE}
}

@article{deng2025collective,
  title={Collective dynamics of particle swarm optimization: A network science perspective},
  author={Deng, Lingyun and Liu, Sanyang},
  journal={Physica A: Statistical Mechanics and its Applications},
  volume={675},
  pages={130778},
  year={2025},
  publisher={Elsevier}
}

@inproceedings{oliveira2013assessing,
  title={Assessing particle swarm optimizers using network science metrics},
  author={Oliveira-J{\'u}nior, Marcos AC and Bastos-Filho, Carmelo JA and Menezes, Ronaldo},
  booktitle={Complex Networks IV: Proceedings of the 4th Workshop on Complex Networks CompleNet 2013},
  pages={173--184},
  year={2013},
  organization={Springer}
}

@article{oliveira2015using,
  title={Using network science to assess particle swarm optimizers},
  author={Oliveira, Marcos and Bastos-Filho, Carmelo JA and Menezes, Ronaldo},
  journal={Social Network Analysis and Mining},
  volume={5},
  number={1},
  pages={3},
  year={2015},
  publisher={Springer}
}

@inproceedings{oliveira2016communication,
  title={Communication diversity in particle swarm optimizers},
  author={Oliveira, Marcos and Pinheiro, Diego and Andrade, Bruno and Bastos-Filho, Carmelo and Menezes, Ronaldo},
  booktitle={International Conference on Swarm Intelligence},
  pages={77--88},
  year={2016},
  organization={Springer}
}

@article{fortunato2007resolution,
  title={Resolution limit in community detection},
  author={Fortunato, Santo and Barthelemy, Marc},
  journal={Proceedings of the national academy of sciences},
  volume={104},
  number={1},
  pages={36--41},
  year={2007},
  publisher={National Academy of Sciences}
}

@article{arenas2008analysis,
  title={Analysis of the structure of complex networks at different resolution levels},
  author={Arenas, Alex and Fernandez, Alberto and Gomez, Sergio},
  journal={New journal of physics},
  volume={10},
  number={5},
  pages={053039},
  year={2008}
}

@article{blondel2008fast,
  title={Fast unfolding of communities in large networks},
  author={Blondel, Vincent D and Guillaume, Jean-Loup and Lambiotte, Renaud and Lefebvre, Etienne},
  journal={Journal of statistical mechanics: theory and experiment},
  volume={2008},
  number={10},
  pages={P10008},
  year={2008}
}

@article{fiedler1973algebraic,
  title={Algebraic connectivity of graphs},
  author={Fiedler, Miroslav},
  journal={Czechoslovak mathematical journal},
  volume={23},
  number={2},
  pages={298--305},
  year={1973},
  publisher={Institute of Mathematics, Academy of Sciences of the Czech Republic}
}

@article{lim2018self,
  title={A self-adaptive topologically connected-based particle swarm optimization},
  author={Lim, Wei Hong and Isa, Nor Ashidi Mat and Tiang, Sew Sun and Tan, Teng Hwang and Natarajan, Elango and Wong, Chin Hong and Tang, Jing Rui},
  journal={IEEE Access},
  volume={6},
  pages={65347--65366},
  year={2018},
  publisher={IEEE}
}

@inproceedings{varna2020hidms,
  title={HIDMS-PSO: A new heterogeneous improved dynamic multi-swarm PSO algorithm},
  author={Varna, Fevzi Tugrul and Husbands, Phil},
  booktitle={2020 IEEE Symposium Series on Computational Intelligence (SSCI)},
  pages={473--480},
  year={2020},
  organization={IEEE}
}

@article{vafashoar2022bifurcated,
  title={Bifurcated particle swarm optimizer with topology learning particles},
  author={Vafashoar, Reza and Morshedlou, Hossein and Meybodi, Mohammad Reza},
  journal={Applied Soft Computing},
  volume={114},
  pages={108039},
  year={2022},
  publisher={Elsevier}
}

@article{wu2023incorporating,
  title={Incorporating surprisingly popular algorithm and euclidean distance-based adaptive topology into PSO},
  author={Wu, Xuan and Han, Jizong and Wang, Di and Gao, Pengyue and Cui, Quanlong and Chen, Liang and Liang, Yanchun and Huang, Han and Lee, Heow Pueh and Miao, Chunyan and others},
  journal={Swarm and Evolutionary Computation},
  volume={76},
  pages={101222},
  year={2023},
  publisher={Elsevier}
}

@article{pei2026hpso,
  title={HPSO: Particle Swarm Optimization with Hypergraph-Based Topology},
  author={Pei, Wenbin and Luo, Xi and Xue, Bing and Zhang, Mengjie and Zhang, Qiang},
  journal={arXiv preprint arXiv:2608.07587},
  year={2026}
}

@article{houssein2021major,
  title={Major advances in particle swarm optimization: theory, analysis, and application},
  author={Houssein, Essam H and Gad, Ahmed G and Hussain, Kashif and Suganthan, Ponnuthurai Nagaratnam},
  journal={Swarm and Evolutionary Computation},
  volume={63},
  pages={100868},
  year={2021},
  publisher={Elsevier}
}

@inproceedings{shi1999empirical,
  title={Empirical study of particle swarm optimization},
  author={Shi, Yuhui and Eberhart, Russell C},
  booktitle={Proceedings of the 1999 congress on evolutionary computation-CEC99 (Cat. No. 99TH8406)},
  volume={3},
  pages={1945--1950},
  year={1999},
  organization={IEEE}
}

@article{ratnaweera2004self,
  title={Self-organizing hierarchical particle swarm optimizer with time-varying acceleration coefficients},
  author={Ratnaweera, Asanga and Halgamuge, Saman K and Watson, Harry C},
  journal={IEEE Transactions on evolutionary computation},
  volume={8},
  number={3},
  pages={240--255},
  year={2004},
  publisher={IEEE}
}

@inproceedings{liang2005dynamic,
  title={Dynamic multi-swarm particle swarm optimizer},
  author={Liang, Jane-Jing and Suganthan, Ponnuthurai Nagaratnam},
  booktitle={Proceedings 2005 IEEE Swarm Intelligence Symposium, 2005. SIS 2005.},
  pages={124--129},
  year={2005},
  organization={IEEE}
}

@article{janson2005hierarchical,
  title={A hierarchical particle swarm optimizer and its adaptive variant},
  author={Janson, Stefan and Middendorf, Martin},
  journal={IEEE Transactions on Systems, Man, and Cybernetics, Part B (Cybernetics)},
  volume={35},
  number={6},
  pages={1272--1282},
  year={2005},
  publisher={IEEE}
}

@article{guo2026adaptive,
  title={An adaptive dynamic multi-swarm PSO algorithm based on topological structure for solving complex constrained optimization problems and its applications on UAV path planning},
  author={Guo, Eryang and Gao, Yuelin and Hu, Chenyang},
  journal={Applied Soft Computing},
  pages={114813},
  year={2026},
  publisher={Elsevier}
}

@article{hao2025q,
  title={Q-Learning-Based Multi-Strategy Topology Particle Swarm Optimization Algorithm},
  author={Hao, Xiaoxi and Wang, Shenwei and Liu, Xiaotong and Wang, Tianlei and Qiu, Guangfan and Zeng, Zhiqiang},
  journal={Algorithms},
  volume={18},
  number={11},
  pages={672},
  year={2025},
  publisher={MDPI}
}

@inproceedings{oliveira2014towards,
  title={Towards a network-based approach to analyze particle swarm optimizers},
  author={Oliveira, Marcos and Bastos-Filho, Carmelo JA and Menezes, Ronaldo},
  booktitle={2014 IEEE Symposium on Swarm Intelligence},
  pages={1--8},
  year={2014},
  organization={IEEE}
}

@inproceedings{yin2025self,
  title={Self-evaluated topology adaptation in particle swarm optimization using Metropolis criterion},
  author={Yin, Kang and Yang, Yihao and Sato, Yuji},
  booktitle={Proceedings of the 2025 9th International Conference on Intelligent Systems, Metaheuristics \& Swarm Intelligence},
  pages={13--19},
  year={2025}
}

@article{yang2023adaptive,
  title={An adaptive dynamic multi-swarm particle swarm optimization with stagnation detection and spatial exclusion for solving continuous optimization problems},
  author={Yang, Xu and Li, Hongru and Huang, Youhe},
  journal={Engineering Applications of Artificial Intelligence},
  volume={123},
  pages={106215},
  year={2023},
  publisher={Elsevier}
}

@inproceedings{ahmed2014efficient,
  title={An efficient fitness-based stagnation detection method for particle swarm optimization},
  author={Ahmed, Hazem Radwan},
  booktitle={Proceedings of the Companion Publication of the 2014 Annual Conference on Genetic and Evolutionary Computation},
  pages={1029--1032},
  year={2014}
}

@article{wu2017problem,
  title={Problem definitions and evaluation criteria for the CEC 2017 competition on constrained real-parameter optimization},
  author={Wu, Guohua and Mallipeddi, Rammohan and Suganthan, Ponnuthurai Nagaratnam},
  journal={National University of Defense Technology, Changsha, Hunan, PR China and Kyungpook National University, Daegu, South Korea and Nanyang Technological University, Singapore, Technical Report},
  volume={9},
  number={10},
  pages={2017},
  year={2017}
}

@misc{mej2010networks,
  title={Networks: an introduction},
  author={Mej, Newman},
  year={2010},
  publisher={Oxford University Press, Oxford}
}

@article{barrat2004architecture,
  title={The architecture of complex weighted networks},
  author={Barrat, Alain and Barthelemy, Marc and Pastor-Satorras, Romualdo and Vespignani, Alessandro},
  journal={Proceedings of the national academy of sciences},
  volume={101},
  number={11},
  pages={3747--3752},
  year={2004},
  publisher={National academy of sciences}
}

@article{clerc2002particle,
  title={The particle swarm - explosion, stability, and convergence in a multidimensional complex space},
  author={Clerc, Maurice and Kennedy, James},
  journal={IEEE Transactions on Evolutionary Computation},
  volume={6},
  number={1},
  pages={58--73},
  year={2002},
}

@techreport{clerc2003tribes,
  title={Tribes, a parameter free particle swarm optimizer},
  author={Clerc, Maurice},
  year={2003},
  note={Available: \url{http://clerc.maurice.free.fr/pso/}}
}

@techreport{clerc2006stagnation,
  title={Stagnation analysis in particle swarm optimisation or what happens when nothing happens},
  author={Clerc, Maurice},
  institution={Department of Computer Science, University of Essex},
  number={CSM-460},
  year={2006},
  note={Available: \url{https://hal.archives-ouvertes.fr/hal-00122031}}
}

@techreport{clerc2006confinements,
  title={Confinements and biases in particle swarm optimization},
  author={Clerc, Maurice},
  year={2006},
  note={Available: \url{http://clerc.maurice.free.fr/pso/}}
}

@article{spearman1904proof,
  author  = {Spearman, Charles},
  title   = {The Proof and Measurement of Association Between Two Things},
  journal = {The American Journal of Psychology},
  year    = {1904},
  volume  = {15},
  number  = {1},
  pages   = {72--101},
}

@book{box2015time,
  author    = {Box, George E. P. and Jenkins, Gwilym M. and Reinsel, Gregory C. and Ljung, Greta M.},
  title     = {Time Series Analysis: Forecasting and Control},
  edition   = {5th},
  publisher = {Wiley},
  address   = {Hoboken, NJ},
  year      = {2015},
  isbn      = {978-1-118-67502-1}
}

@article{prichard1995generalized,
  author  = {Prichard, Dean and Theiler, James},
  title   = {Generalized Redundancies for Time Series Analysis},
  journal = {Physica D: Nonlinear Phenomena},
  year    = {1995},
  volume  = {84},
  number  = {3--4},
  pages   = {476--493},
}

@article{valencia2023influence,
  title={Influence of the number of connections between particles in the performance of a multi-objective particle swarm optimizer},
  author={Valencia-Rodr{\'\i}guez, Diana Cristina and Coello, Carlos A Coello},
  journal={Swarm and Evolutionary Computation},
  volume={77},
  pages={101231},
  year={2023},
  publisher={Elsevier}
}

@article{yang2023random,
  title={Random contrastive interaction for particle swarm optimization in high-dimensional environment},
  author={Yang, Qiang and Song, Gong-Wei and Chen, Wei-Neng and Jia, Ya-Hui and Gao, Xu-Dong and Lu, Zhen-Yu and Jeon, Sang-Woon and Zhang, Jun},
  journal={IEEE Transactions on Evolutionary Computation},
  volume={28},
  number={4},
  pages={933--949},
  year={2023},
  publisher={IEEE}
}

@inproceedings{gupta2025enhancing,
  title={Enhancing explainability and reliable decision-making in particle swarm optimization through communication topologies},
  author={Gupta, Nitin and Bala, Indu and Dutta, Bapi and Mart{\'\i}nez, Luis and Yadav, Anupam},
  booktitle={Proceedings of the Genetic and Evolutionary Computation Conference Companion},
  pages={955--958},
  year={2025}
}
\end{document}